\documentclass[final]{cvpr}

\usepackage{times}
\usepackage{epsfig}
\usepackage{graphicx}
\usepackage{amsmath}
\usepackage{amssymb}
\usepackage{algorithm2e}
\usepackage{multirow}
\usepackage{comment}
\makeatletter
\@namedef{ver@everyshi.sty}{}
\makeatother
\usepackage{tikz}
\usetikzlibrary{shapes,arrows}
\usetikzlibrary{decorations.pathmorphing,decorations.pathreplacing}
\usetikzlibrary{backgrounds,positioning,fit,petri}
\usepackage{color}
\usepackage{xcolor}
\usepackage{mathtools, nccmath}
\usepackage{pifont}
\usepackage{hhline}
\usepackage{makecell}
\usepackage{bm}
\usepackage[percent]{overpic}
\usepackage[outline]{contour}
\usepackage[super]{nth}

\newcommand\blfootnote[1]{%
	\begingroup
	\renewcommand\thefootnote{}\footnote{#1}%
	\addtocounter{footnote}{-1}%
	\endgroup
}

\newcommand{\cmark}{\ding{51}}%
\newcommand{\xmark}{\ding{55}}%

\usepackage[pagebackref=true,breaklinks=true,colorlinks,bookmarks=false]{hyperref}

\def\cvprPaperID{1786} 
\def\confYear{CVPR 2021}

\begin{document}
	\bibliographystyle{unsrt}
	
	\title{Modular Interactive Video Object Segmentation: \\ Interaction-to-Mask, Propagation and Difference-Aware Fusion}
	
	\author{
		Ho Kei Cheng\\
		UIUC/HKUST\\
		{\tt\small hokeikc2@illinois.edu}
		\and
		Yu-Wing Tai\\
		Kuaishou Technology\\
		{\tt\small yuwing@gmail.com}
		\and
		Chi-Keung Tang\\
		HKUST\\
		{\tt\small cktang@cs.ust.hk}
	}
	
	\maketitle
	
	\begin{abstract}
		We present Modular interactive VOS (MiVOS) framework which decouples interaction-to-mask and mask propagation, allowing for higher generalizability and better performance. Trained separately, the interaction module converts user interactions to an object mask, which is then temporally propagated by our 
		propagation module using a novel top-$k$ filtering strategy in reading the space-time memory. To effectively take the user's intent into account, a novel difference-aware module is proposed to learn how to properly fuse the masks before and after each interaction, which are aligned with the target frames by  employing the space-time memory.
		We evaluate our method both qualitatively and quantitatively with different forms of user interactions (e.g., scribbles, clicks) on DAVIS to show that our method outperforms current state-of-the-art algorithms while requiring fewer frame interactions, with the additional advantage in generalizing to different types of user interactions.
		We contribute a large-scale synthetic VOS dataset with pixel-accurate segmentation of 4.8M frames to accompany our source codes to facilitate future research.
	\end{abstract}
	
	\vspace{-2.1em}
	
	\blfootnote{Source code, pretrained models and dataset are available at:  {\url{https://hkchengrex.github.io/MiVOS}}. This research is supported in part by Kuaishou Technology and the Research Grant Council of the Hong Kong SAR under grant no. 16201420.}
	
	\section{Introduction}
	Video object segmentation (VOS) aims to produce high-quality segmentation of a target object instance across an input video sequence, which has wide applications in video understanding and editing. 
	Existing VOS methods can be categorized by the types of user input: semi-supervised methods require pixel-wise annotation of the first frame, while interactive VOS approaches take user interactions (e.g., scribbles or clicks) as input where users can iteratively refine the results  until satisfaction. 
	
	This paper focuses on interactive VOS (iVOS) which finds more applications in video editing, because 
	typical user interactions such as scribbles or clicks (a few seconds per frame) are much easier than specifying full annotation ($\sim$79 seconds per instance), with the iterative or successive refinement scheme allowing the user more control over result accuracy versus interaction budget trade-off~\cite{Caelles_arXiv_2018}.
	
	\begin{figure}[t]
		\begin{center}
			\includegraphics[width=\linewidth]{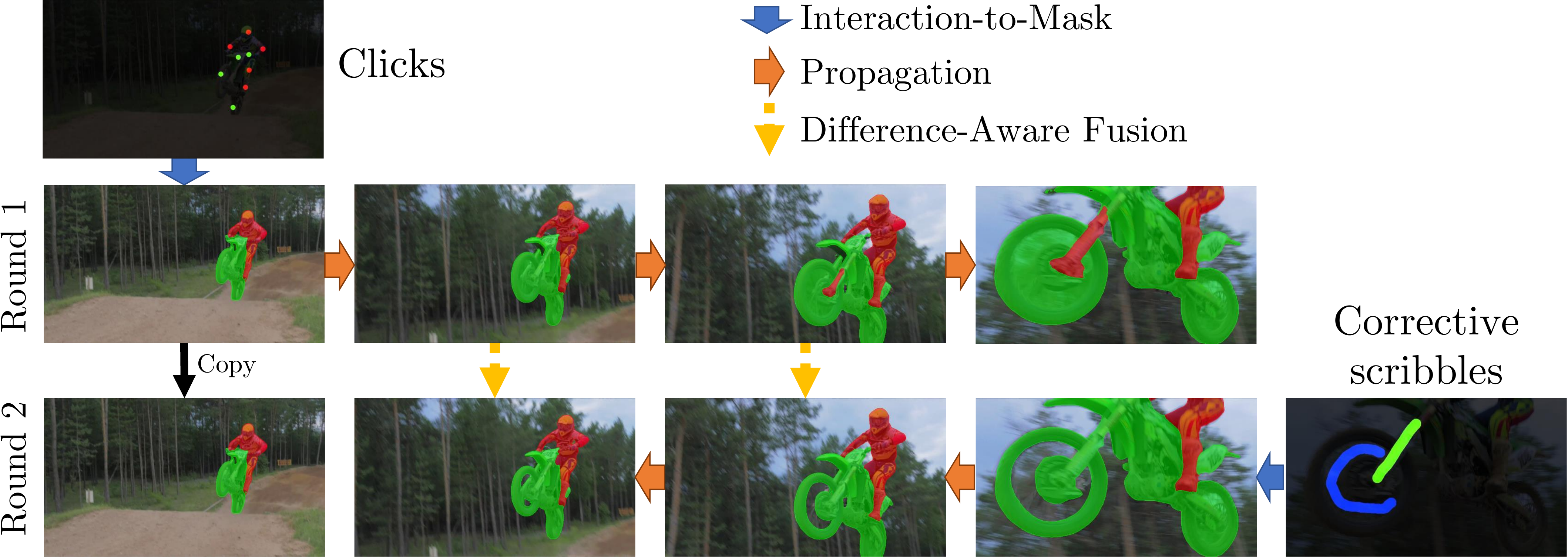}
		\end{center}
		\vspace{-0.15in}
		\caption{
			User annotates one of the frames (e.g.,\ with clicks at the top-left frame) and MiVOS bidirectionally propagates the masks to the entire video sequence. 
			Our difference-aware fusion module guides the segmentation network to
			correct the masks across frames based on user's intended correction on another frame (e.g., with scribbles on the bottom-right frame).}
		\label{fig:teaser}
		\vspace{-0.2in}
	\end{figure}
	
	Conceptually, iVOS can be considered as the combination of two tasks: interaction understanding (e.g.,\ mask generation from interactions~\cite{rother2004grabcut, xu2017deepGrabCut, sofiiuk2020fbrs, maninis2018deepExtremePointsCut}) and temporal propagation (e.g.,\ semi-supervised VOS methods \cite{perazzi2017learningMaskTrack, oh2019videoSTM, luiten2018premvos}). Current methods usually perform the two tasks jointly, using interconnected encoders~\cite{oh2019fastInteractive, heo2019interactiveSDINet, Yuk2020IVOSGlobalLocal} or memory-augmented interaction features~\cite{miao2020memoryAggregationInteractive, Chen2020WorkshopCFBI, oh2020STMPAMI}. The strong coupling limits the form of user interaction (e.g.,\ scribbles only) and makes training difficult. Attempts to decouple the two tasks fail to reach state-of-the-art accuracy~\cite{benard2017interactiveInTheWild, Tran2020WorkshopMultiple} as user's intent cannot be adequately taken into account in the propagation process.
	
	One advantage of unified methods over decoupled methods is that the former can efficiently pick up small corrective interactions across many frames, which is suited to the DAVIS evaluation robot~\cite{Caelles_arXiv_2018}. 
	However, we believe that human users tend to interactively correct a single frame to high accuracy before checking other frames, as the visual examination itself takes time and human labor while free for an evaluation robot. 
	Our method requires less interacted frames by letting the user focus on a single frame multiple times while attaining the same or even better accuracy. Our method is efficient as single-frame interaction can be done almost instantly~\cite{sofiiuk2020fbrs}, with the more time-consuming propagation performed only sparsely.
	
	In this paper we present a decoupled modular framework to address the iVOS problem. Note that na\"ive decoupling may lead to loss of user's intent as the original interaction is no longer available in the propagation stage. This problem is circumvented by our new difference-aware fusion module which models the difference in the mask before and after each interaction to inject the user's intent in propagation. Thus
	the user's intent is preserved and propagated to the rest of the video sequence.
	We argue that  \emph{mask difference} is a better representation than raw interactions which is unambiguous and does not depend on interaction types. With our decoupling approach, our method can accept different types of user interactions and achieve better performance on various qualitative and quantitative evaluations.
	Our main contributions can be summarized as follows:
	\begin{itemize}
		\vspace{-5pt}
		\item We innovate on the decoupled interaction-propagation framework and show that this approach is simple, effective, and generalizable.
		\vspace{-5pt}
		\item 
		We propose a novel lightweight top-$k$ filtering scheme for the attention-based memory read operation in mask generation during propagation.
		\vspace{-5pt}
		\item We propose a novel difference-aware fusion module to faithfully capture the user's intent which improves iVOS accuracy and reduces the amount of user interaction. 
		We will show how to efficiently align the masks before and after an  interaction  at the target frames by using the space-time memory in propagation.
		\vspace{-5pt}
		\item We contribute a large-scale synthetic VOS dataset with 4.8M frames to accompany our source codes to facilitate future research.
	\end{itemize}
	
	\section{Related Works}
	
\begin{figure}[t]
		\begin{center}
			\includegraphics[width=\linewidth]{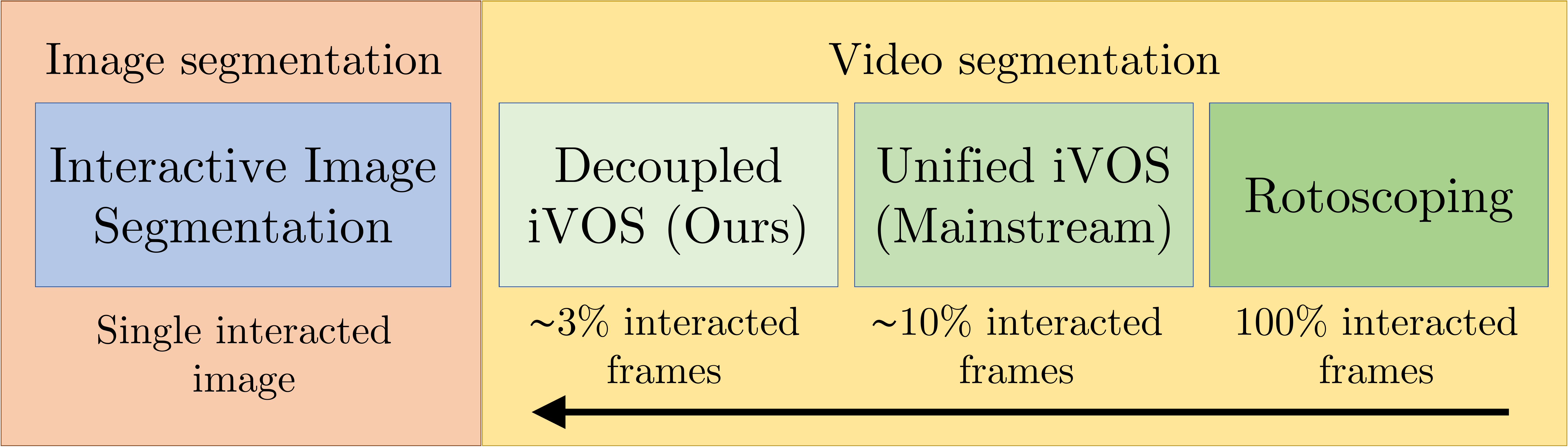}
		\end{center}
		\vspace{-0.15in}
		\caption{
			Progress in iVOS~\cite{Caelles_arXiv_2019} has significantly reduced the amount of human labor required to segment objects in videos compared with traditional rotoscoping methods. By leveraging more spatially dense yet temporally sparse interactions, our method further reduces the human effort required to examine the output video in a more tedious, back-and-forth manner (see Section~\ref{user_study} for user study) while reaching the same or even better accuracy. Our method can be regarded as lifting 2D image segmentation to 3D.
		}
		\label{fig:motivation}
		\vspace{-0.15in}
	\end{figure}

	Figure~\ref{fig:motivation} positions our MiVOS with other related works in interactive image/video object segmentation.
	
	\noindent\textbf{Semi-Supervised Video Object Segmentation}. 
	This task aims to segment a specific object throughout a video given only a fully-annotated mask in the first frame.
	Early methods often employ test-time finetuning on the given frame~\cite{luiten2018premvos, voigtlaender2017onlineOnAVOS, caelles2017oneOSVOS, khoreva2017lucid, perazzi2017learningMaskTrack, xu2019spatiotemporal} to improve the model's discriminatory power, but such finetuning is  often too slow. 
	Recently, diverse approaches have been explored including pixel-wise embedding~\cite{voigtlaender2019feelvos, chen2018blazinglyFast, yang2020collaborativeCFBI}, mask propagation and tracking~\cite{perazzi2017learningMaskTrack, hu2017maskrnn, ventura2019rvos, wang2019ranet, yang2018efficientModulation, oh2018fastRGMP, wang2019fastSiamMask, chen2020stateAwareTracker, cheng2018fastTrackingParts}, building a target model~\cite{robinson2020learningTargetModel}, and memory features matching~\cite{hu2018videomatch, oh2019videoSTM, li2020fastGlobalContext, miao2020memoryAggregationInteractive, huang2020fastTemporalAggregation, Wen2020DMVOS, Duarte2019Capsule}. 
	In particular, STM~\cite{oh2019videoSTM} constructs a memory bank from past frames and predicts the mask using a query-key-value attention mechanism. While simple and effective, this method can achieve state-of-the-art results.
	In this work, we propose to transfer the technical progress of semi-supervised VOS methods to the interactive domain. Our space-time memory network, which is inspired by STM~\cite{oh2019videoSTM}, is used in our propagation backbone.
	
	\noindent\textbf{Interactive Video Object Segmentation (iVOS)}.
	User-supplied hints are provided in iVOS. The interactions can be used to either segment an object or a correct previously misclassified region~\cite{wang2005interactiveVideoCutout, price2009livecut, shankar2015video, Caelles_arXiv_2018}. Most recent works~\cite{Yuk2020IVOSGlobalLocal, oh2019fastInteractive, miao2020memoryAggregationInteractive} have focused on scribble interaction which is used and provided by the DAVIS challenge~\cite{Caelles_arXiv_2019}.
	A recent method~\cite{chen2018blazinglyFast} has extended their embedding network in the interactive setting with clicks as user input.
	Our method can generalize to a wide range of user interactions due to the modular design by simply replacing the interaction-to-mask component.
	
	\begin{figure*}[t]
		\includegraphics[width=\linewidth]{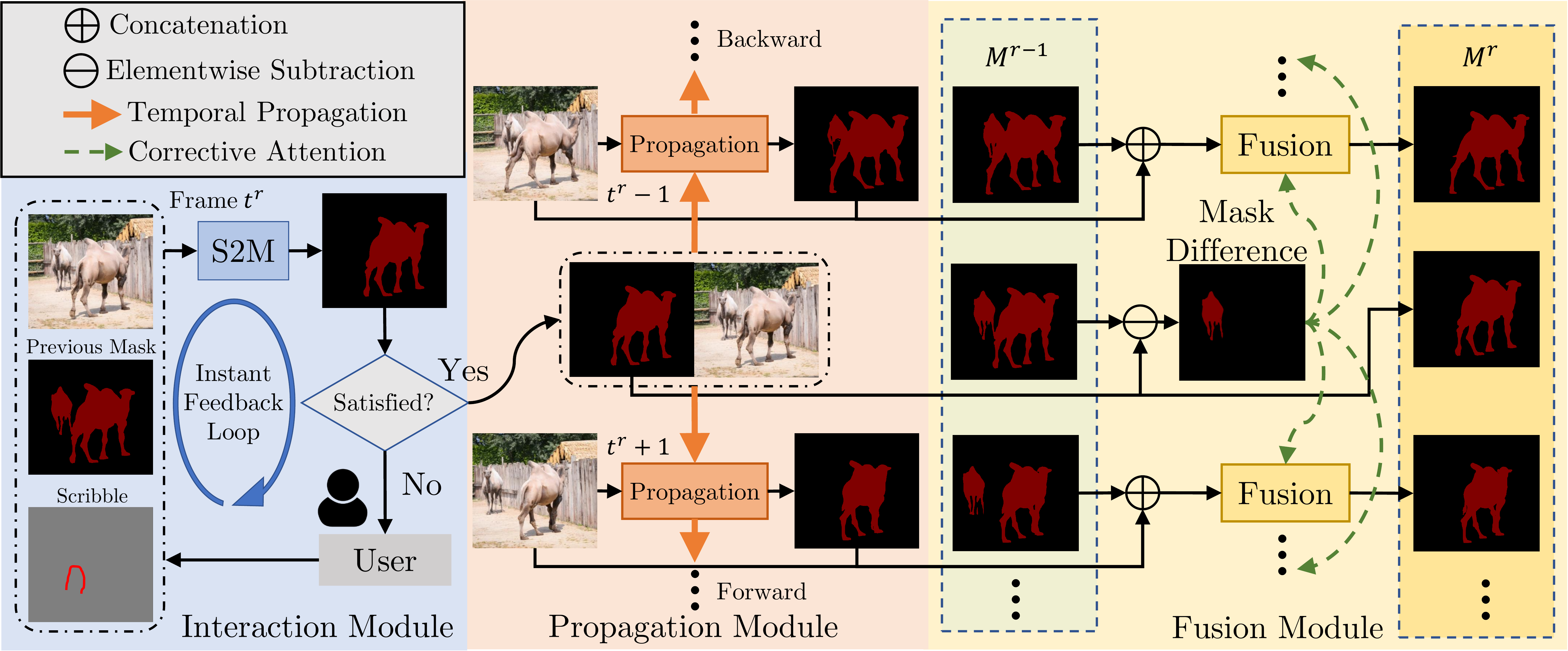}
		\vspace{-0.15in}
		\caption{\textbf{MiNet} overview. 
			In interaction round $r$, the user picks a frame $t'$ and interactively correct the object mask 
			until satisfaction using the Scribble-to-Mask~(S2M) module~(Section~\ref{i2m}) running in real time. 
			The corrected mask will then be bidirectionally propagated through the video sequence with the propagation module~(Section~\ref{prop}).
			To incorporate information from previous rounds, a difference-aware fusion module is used 
			to fuse previous and current masks. The difference in the interacted mask before and after the interaction (which conveys user's intention) is used in the fusion module via an attention mechanism~(Section~\ref{fusion}). In the first round, all masks are initialized to zeros.}
		\label{fig:overview}
		\vspace{-0.15in}
	\end{figure*}
	
	The majority of current deep learning based iVOS methods is based on deep feature fusion to incorporate user interactions into the segmentation task, where two interconnected encoder networks are designed~\cite{oh2019fastInteractive, heo2019interactiveSDINet,Yuk2020IVOSGlobalLocal}, or scribble features are stored as memory which are referenced later in the segmentation process~\cite{miao2020memoryAggregationInteractive, Chen2020WorkshopCFBI, oh2020STMPAMI}. These approaches inevitably tie the particular form of user inputs with the mask propagation process. This property makes training difficult as the model needs to adapt to both understanding the interactions and accurately propagating masks at the same time. 
	Alternatively, some methods have attempted to decouple the interaction and propagation network~\cite{benard2017interactiveInTheWild, Tran2020WorkshopMultiple} by first generating a mask given an interaction in any types, followed by propagating this mask bidirectionally. But these methods fail to achieve state-of-the-art performance. We believe that this is due to the dismissal of user intent as the propagation network no longer has access to the original user interaction.
	
	This paper proposes to overcome the above problem by considering the difference in the mask domain before and after an interaction round in order to directly and faithfully represent the user intent in the propagation process.
	
	\noindent\textbf{Interactive Image Segmentation}.
	The problem of interactive image segmentation or cutout has a long history with a wide range of applications~\cite{li2004lazy, mortensen1995intelligent, kass1988snakes, rother2004grabcut}. The recent adoption of deep convolutional neural network has greatly improved state-of-the-art performance with different types of user interactions such as bounding boxes~\cite{xu2017deepGrabCut}, clicks~\cite{xu2016deepObjectSelection, sofiiuk2020fbrs, sofiiuk2020fbrs}, or extreme points~\cite{maninis2018deepExtremePointsCut, agustsson2019interactiveAllRegions}.
	Our modular approach can adapt to any of these types of interactions by adopting the corresponding interaction-to-mask algorithm in our framework.
	\vspace{-4pt}
	
	\section{Method}
	\vspace{-1pt}
	Initially, the user selects and interactively annotates one frame (e.g.,~using scribbles or clicks) to produce a mask. Our method then generates segmentation for every frame in the video sequence. After that, the user examines the output quality, and if needed, starts a new ``round'' by correcting an erroneous frame with further interactions. 
	We denote $r$ as the current interaction round. Using superscript, the user-interacted frame index in the $r$-th round is $t^r$, and the mask results of the $r$-th round is $M^r$; using subscript, the mask of individual $j$-th frame is denoted as $M^r_j$. Refer to supplementary material for a quick index of the paper's notations.
	
	\vspace{-2pt}
	\subsection{MiNet Overview}
	\vspace{-1pt}
	As illustrated in Figure~\ref{fig:overview}, our method consists of three core components: interaction-to-mask, mask propagation, and difference-aware fusion. 
	The interaction module operates in an instant feedback loop, allowing the user to obtain real-time feedback and achieve a satisfactory result on a single frame before the more time-consuming propagation process\footnote{To the best of our knowledge, most related state-of-the-art works take $>100$ms per frame, with current ``fast" methods taking $>15$ms per frame for propagation. This justifies our single-frame interaction and propagation where the latter runs at $\sim100$ms per frame}.
	In the propagation module, the corrected mask is bidirectionally propagated independently of $M^{r-1}$. 
	Finally, the propagated masks are fused with $M^{r-1}$ with the fusion module which aims to fuse the two sequences while avoiding possible decay or loss of user's intent. The user intent is captured using the difference in the selected mask before and after user interaction. This difference is fed into the fusion module as guidance.
	
	\subsection{Interaction-to-Mask}\label{i2m}
	Various interactive image segmentation methods can be used here as long as they can compute an object mask from user interactions.
	Users are free to use their favorite segmentation tool or even tailored pipeline for specific tasks (e.g.,\ human segmentation for movie editing). 
	Methods that use information from an existing mask ($M^{r-1}_{t^r}$) might be more labor-efficient but such property is optional.
	
	We design a Scribble-to-Mask~(S2M) network to evaluate our method on the DAVIS~\cite{Caelles_arXiv_2019} benchmark. Our pipeline has high versatility  not restricted by any one type of such interaction network -- we additionally employ click-based interaction~\cite{sofiiuk2020fbrs}, freehand drawing, and a local control module that allows fine adjustment which are experimented in the user study Section~\ref{user_study}.
	
	\vspace{0.1in}
	\noindent\textbf{S2M}\space\space
	The goal of the S2M network is to produce a single-image segmentation in real time given input scribbles.
	Our design is intentionally straightforward with a standard DeepLabV3+~\cite{chen2018encoderDeepLabV3+} semantic segmentation network as the backbone. The network takes a six-channel input: RGB image, existing mask, and positive/negative scribble maps, and deals with two cases: initial interaction (where the existing mask is empty) and corrective interaction (where the existing mask contains error).
	Unlike previous methods~\cite{oh2020STMPAMI, oh2019fastInteractive, Yuk2020IVOSGlobalLocal}, we train with a simpler single-round approach on a large collection of static images~\cite{gupta2019lvis, zeng2019towardsHRSOD, shi2015hierarchicalECSSD, FSS1000}. We are able to leverage these non-video large datasets by the virtue of our decoupled paradigm.
	
	For each input image, we randomly pick one of the two cases (with an empirically set probability of $0.5$) and synthesize the corresponding input mask which is either set to zeros or perturbed from the ground-truth with random dilation/erosion~\cite{CascadePSP2020}. We do not reuse the output mask to form a second training stage~\cite{oh2020STMPAMI, oh2019fastInteractive, Yuk2020IVOSGlobalLocal} to reduce training cost and complications. Input scribbles are then generated correspondingly in the error regions using strategies~\cite{Caelles_arXiv_2019} such as thinning or random B\'ezier curves.
	
\begin{figure}[h]
		\vspace{-0.15in}
		\begin{center}
			\includegraphics[width=0.9\linewidth]{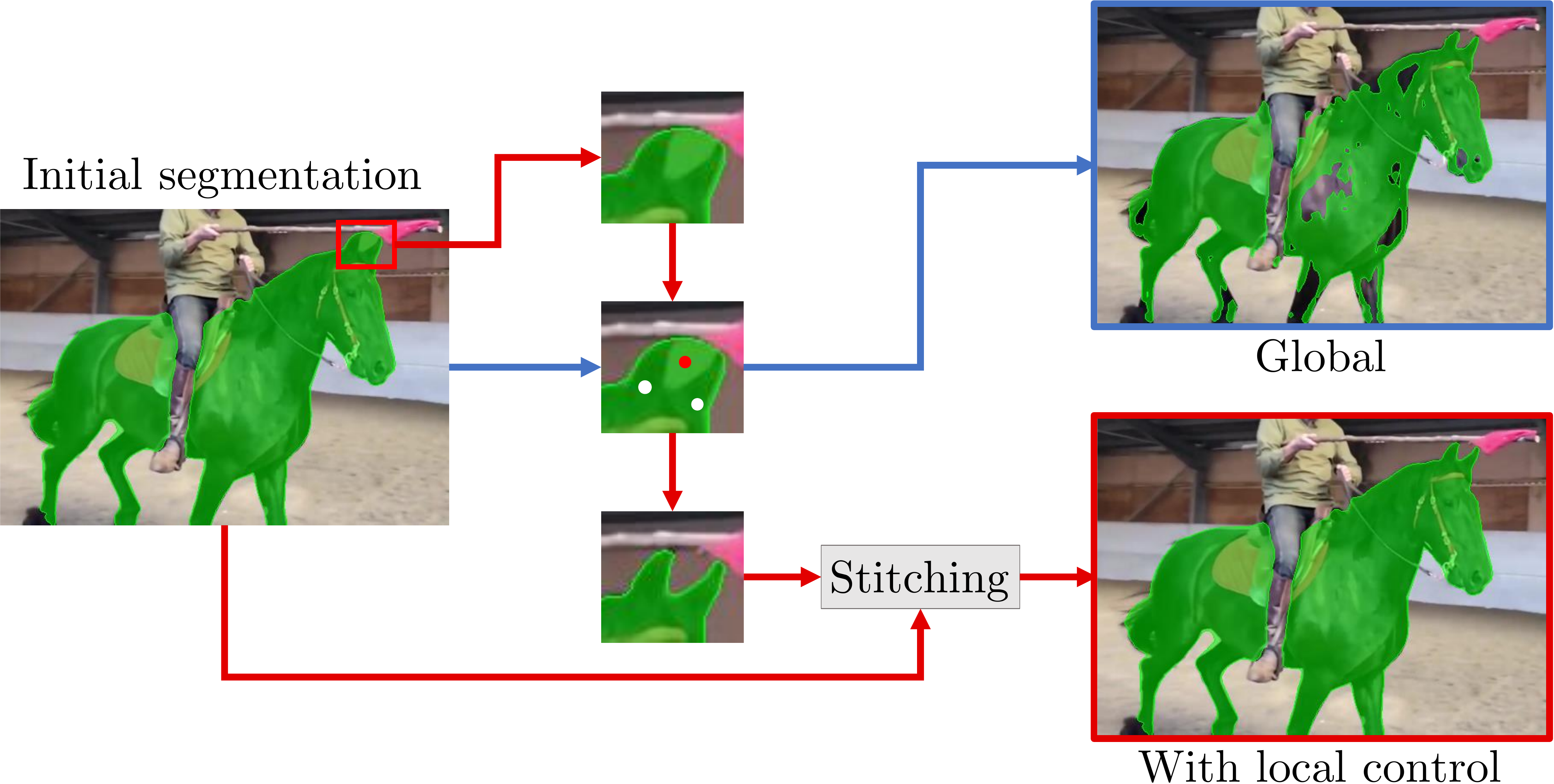}
		\end{center}
		\vspace{-0.15in}
		\caption{
			The local control pathway (red) uses an ROI to prevent deterioration spread by the global interaction path (blue) when only a small local refinement (around ears) is needed.
		}
		\label{fig:global_local}
		\vspace{-0.15in}
	\end{figure}

	\vspace{0.1in}
	\noindent\textbf{Local Control}\space\space
	While state-of-the-art interactive segmentation methods such as f-BRS~\cite{sofiiuk2020fbrs} often use a large receptive field to enable fast segmentation with few clicks, it may harm the global result when only local fine adjustment is needed toward the end of the segmentation process. Figure~\ref{fig:global_local} illustrates one such case where the global shape is correct except for the ears. With our decoupled approach, it is straightforward to assert local control by limiting the interactive algorithm to apply in a user-specified region as shown in the figure. The region's result can be effortlessly stitched back to the main segmentation. 
	
	\subsection{Temporal Propagation}\label{prop}
	
	\begin{figure}[t]
		\begin{center}
			\includegraphics[width=0.9\linewidth]{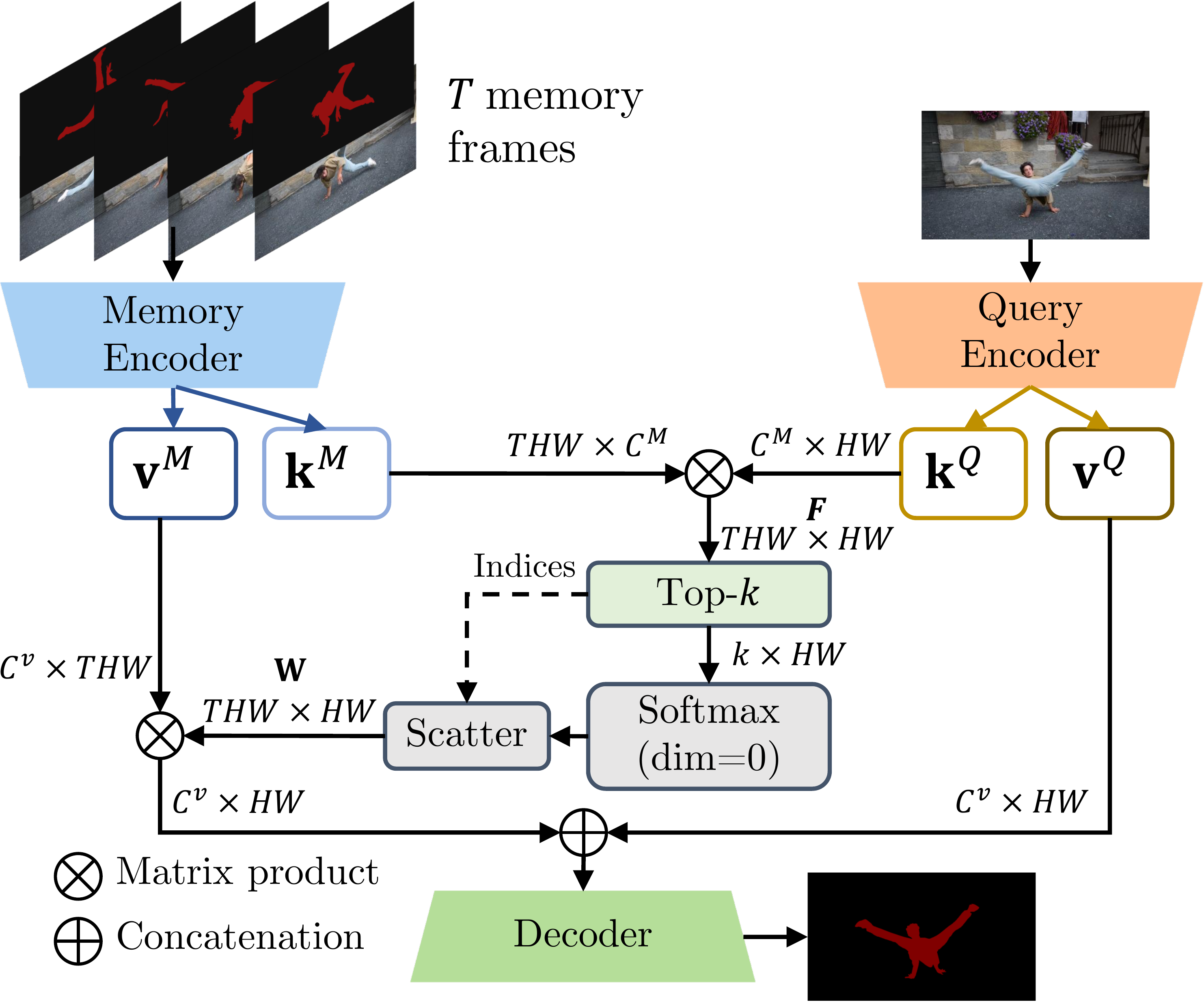}
		\end{center}
		\vspace{-0.15in}
		\caption{
			Implementation of our space-time memory reader as described in Section~\ref{prop}. Tensor reshaping is performed when needed. Skip-connections from the query encoder to the decoder are omitted for clarity. }
		\label{fig:spacetime}
		\vspace{-0.15in}
	\end{figure}
	Given an object mask, the propagation module tracks the object and produces corresponding masks in subsequent frames. Following STM~\cite{oh2019videoSTM}, we consider the past frames with object masks as \textit{memory} frames which are used to predict the object mask for the current  (\textit{query}) frame using an attention-based memory read operation. Notably, we propose a novel and lightweight top-$k$ operation that integrates with STM and show that it improves both performance and speed without complicated training tricks.

	\paragraph{Memory Read with Top-$k$ Filtering}
	We build two encoder networks: the memory encoder and the query encoder. Their network backbones are extracted from ResNet50~\cite{he2016deepResNet} up to stage-4 ({\tt res4}) with a stride of 16. 
	Extra input channels are appended to the first convolution of the memory encoder which accepts object masks as input. 
	At the end of each encoder, two separate convolutions are used to produce two features maps: key $\mathbf{k}\in \mathbb{R}^{C^k\times HW}$ and value $\mathbf{v}\in \mathbb{R}^{C^v\times HW}$ where $H$ and $W$ are the image dimensions after stride, and $C^k$ and $C^v$ are set to 128 and 512 respectively.

	Figure~\ref{fig:spacetime} illustrates our space-time memory read operation. For each of the $T$ memory frames, we compute key-value features and concatenate the output as memory key $\mathbf{k}^M\in \mathbb{R}^{C^k\times THW}$ and memory value $\mathbf{v}^M\in \mathbb{R}^{C^v\times THW}$. The key $\mathbf{k}^Q$ computed from the query is matched with $\mathbf{k}^M$ via a dot product:
	\begin{equation}
		\mathbf{F} = \left(\mathbf{k}^M\right)^T \mathbf{k}^Q, 
	\end{equation}
	where each entry in $\mathbf{F}\in\mathbb{R}^{THW\times HW}$ represents the affinity between a query position and a memory position. Previous methods~\cite{oh2019videoSTM, seong2020kernelizedMemory} would then apply softmax along the memory dimension and use the resultant probability distribution as a weighted-sum for $\mathbf{v^M}$. 
	We have two observations on this softmax strategy: 1)\ For each query position, most of the weights will fall into a small set of memory positions and the rest are noises, and 2)\ these noises grow with the size of the memory and are performance-degrading when the sequence is long. 
	
	\begin{figure}[t]
		\begin{center}
			\includegraphics[width=\linewidth]{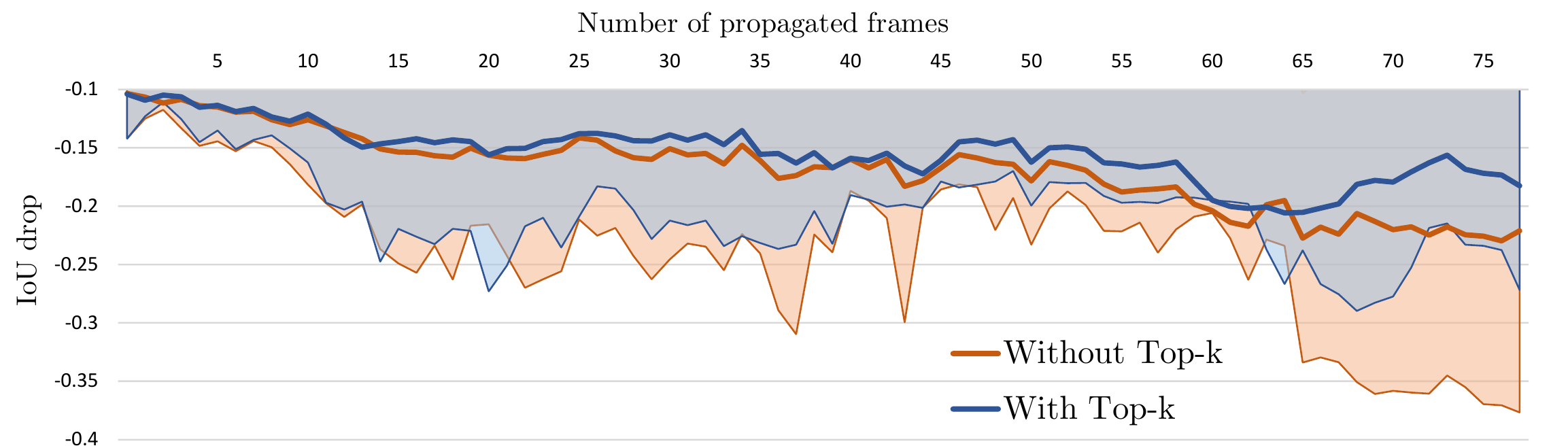}
		\end{center}
		\vspace{-0.1in}
		\caption{
			Mean IoU drop along propagation with or without top\protect\nobreakdash-$k$ filtering with the color bands showing the interquartile range (the higher the better). With top\protect\nobreakdash-$k$ filtering, the propagation is more stable and performs better especially for temporally far away frames when the noise-to-$k$ ratio is large.
		}
		\label{fig:topk}
		\vspace{-0.15in}
	\end{figure}
	
	Based on these observations, we propose to filter the affinities such that only the top-$k$ entries are kept. 
	This effectively removes noises regardless of the sequence length. 
	Since softmax preserves order, we can apply top-$k$ filtering beforehand to reduce the number of expensive $\exp$ calls. 
	In practice, our new top-$k$ strategy not only increases robustness but also overcomes the overhead of top-$k$ (see Table~\ref{tab:run_time}). Figure~\ref{fig:topk} reports the performance increase and robustness brought by top-$k$ filtering.
	Note that KMN~\cite{seong2020kernelizedMemory} (a recent modification of STM) imposes a Gaussian locality prior on the \emph{query} using the \emph{memory}, while our top-$k$ operation filters the \emph{memory} using the \emph{query}. Refer to the supplementary material for a detailed comparison.
	
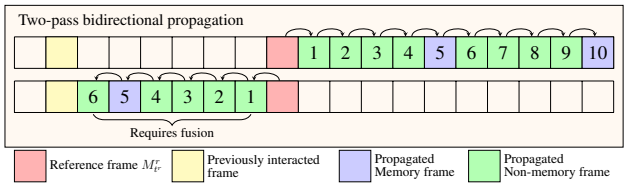
\begin{figure}[h]
		\begin{center}
			\resizebox{\columnwidth}{!}{
				\begin{tikzpicture}
	\tikzstyle{number}=[minimum width=1cm, minimum height=1cm, align=center, font=\LARGE]
	\tikzstyle{prop}=[->, thick,>=stealth]
	\tikzstyle{jump}=[->, ultra thick, dotted,>=stealth]

    	\fill[orange!5!white, draw=black] (1.7,2) rectangle (21.3, -2.5);
    	\node[anchor=west] at (2, 1.5) {\Large Two-pass bidirectional propagation};
    	
    	\fill[red!30!white, draw=black] (2.0,-2.6) rectangle (3.0,-3.6);
    	\node[anchor=west] at (3.0,-3.1) {\large Reference frame $M^r_{t^r}$};
    	
    	\fill[yellow!30!white, draw=black] (7.0,-2.6) rectangle (8.0,-3.6);
    	\node[anchor=west, align=left] at (8.0,-3.1) {\large Previously interacted \\ \large frame};
    	
    	\fill[blue!20!white, draw=black] (12.3,-2.6) rectangle (13.3,-3.6);
    	\node[anchor=west, align=left] at (13.3,-3.1) {\large Propagated \\ \large Memory frame};
    	
    	\fill[green!30!white, draw=black] (16.4,-2.6) rectangle (17.4,-3.6);
    	\node[anchor=west, align=left] at (17.4,-3.1) {\large Propagated \\ \large Non-memory frame};
    	
	
	\draw[step=1cm,black] (2,0) grid (21,1);
	\fill[yellow!30!white, draw=black] (3,0) rectangle (4,1);
	\fill[red!30!white, draw=black] (10,0) rectangle (11,1);
	\fill[green!30!white, draw=black] (11,0) rectangle (12,1);
	\fill[green!30!white, draw=black] (12,0) rectangle (13,1);
	\fill[green!30!white, draw=black] (13,0) rectangle (14,1);
	\fill[green!30!white, draw=black] (14,0) rectangle (15,1);
	\fill[blue!20!white, draw=black] (15,0) rectangle (16,1);
	\fill[green!30!white, draw=black] (16,0) rectangle (17,1);
	\fill[green!30!white, draw=black] (17,0) rectangle (18,1);
	\fill[green!30!white, draw=black] (18,0) rectangle (19,1);
	\fill[green!30!white, draw=black] (19,0) rectangle (20,1);
	\fill[blue!20!white, draw=black] (20,0) rectangle (21,1);
	
	\node[number] at (10.5, 0.5) (b0) {};
	\node[number] at (11.5, 0.5) (b1) {1};
	\node[number] at (12.5, 0.5) (b2) {2};
	\node[number] at (13.5, 0.5) (b3) {3};
	\node[number] at (14.5, 0.5) (b4) {4};
	\node[number] at (15.5, 0.5) (b5) {5};
	\node[number] at (16.5, 0.5) (b6) {6};
	\node[number] at (17.5, 0.5) (b7) {7};
	\node[number] at (18.5, 0.5) (b8) {8};
	\node[number] at (19.5, 0.5) (b9) {9};
	\node[number] at (20.5, 0.5) (b10) {10};
	
	\draw [prop] ([xshift=.1cm]b0.north) to [bend left=90] ([xshift=-.1cm]b1.north);
	\draw [prop] ([xshift=.1cm]b1.north) to [bend left=90] ([xshift=-.1cm]b2.north);
	\draw [prop] ([xshift=.1cm]b2.north) to [bend left=90] ([xshift=-.1cm]b3.north);
	\draw [prop] ([xshift=.1cm]b3.north) to [bend left=90] ([xshift=-.1cm]b4.north);
	\draw [prop] ([xshift=.1cm]b4.north) to [bend left=90] ([xshift=-.1cm]b5.north);
	\draw [prop] ([xshift=.1cm]b5.north) to [bend left=90] ([xshift=-.1cm]b6.north);
	\draw [prop] ([xshift=.1cm]b6.north) to [bend left=90] ([xshift=-.1cm]b7.north);
	\draw [prop] ([xshift=.1cm]b7.north) to [bend left=90] ([xshift=-.1cm]b8.north);
	\draw [prop] ([xshift=.1cm]b8.north) to [bend left=90] ([xshift=-.1cm]b9.north);
	\draw [prop] ([xshift=.1cm]b9.north) to [bend left=90] ([xshift=-.1cm]b10.north);

	\draw[step=1cm,black,yshift=0.6cm] (2,-1) grid (21,-2);
	\fill[yellow!30!white, draw=black] (3,-0.4) rectangle (4,-1.4);
	\fill[green!30!white, draw=black] (4,-0.4) rectangle (5,-1.4);
	\fill[blue!20!white, draw=black] (5,-0.4) rectangle (6,-1.4);
	\fill[green!30!white, draw=black] (6,-0.4) rectangle (7,-1.4);
	\fill[green!30!white, draw=black] (7,-0.4) rectangle (8,-1.4);
	\fill[green!30!white, draw=black] (8,-0.4) rectangle (9,-1.4);
	\fill[green!30!white, draw=black] (9,-0.4) rectangle (10,-1.4);
	\fill[red!30!white, draw=black] (10,-0.4) rectangle (11,-1.4);
	
	\node[number] at (10.5, -0.9) (a0) {};
	\node[number] at (9.5, -0.9) (a1) {1};
	\node[number] at (8.5, -0.9) (a2) {2};
	\node[number] at (7.5, -0.9) (a3) {3};
	\node[number] at (6.5, -0.9) (a4) {4};
	\node[number] at (5.5, -0.9) (a5) {5};
	\node[number] at (4.5, -0.9) (a6) {6};
	
	\draw [prop] ([xshift=-.1cm]a0.north) to [bend right=90] ([xshift=.1cm]a1.north);
	\draw [prop] ([xshift=-.1cm]a1.north) to [bend right=90] ([xshift=.1cm]a2.north);
	\draw [prop] ([xshift=-.1cm]a2.north) to [bend right=90] ([xshift=.1cm]a3.north);
	\draw [prop] ([xshift=-.1cm]a3.north) to [bend right=90] ([xshift=.1cm]a4.north);
	\draw [prop] ([xshift=-.1cm]a4.north) to [bend right=90] ([xshift=.1cm]a5.north);
	\draw [prop] ([xshift=-.1cm]a5.north) to [bend right=90] ([xshift=.1cm]a6.north);
	
	\draw [thick,decorate,decoration={brace,amplitude=10pt,raise=0.3ex}]
  		(a1.south) -- (a6.south) node[midway,below,yshift=-1em]{\large Requires fusion};
	
\end{tikzpicture}
			}
		\end{center}
		\vspace{-0.1in}
		\caption{
			Illustration of our propagation scheme. The frames between the current reference frame and previously interacted frame require fusion which is described in Section~\ref{fusion}.
		}
		\label{fig:propagation}
		\vspace{-0.1in}
	\end{figure}

	In summary, the affinity of memory position~$i$ with query position~$j$ can be computed by:
	\begin{equation}
		\mathbf{W}_{ij} = \frac{\exp\left( \mathbf{F}_{ij} \right)}{\sum_{p\in\text{Top}^k_j(\mathbf{F})}\left(\exp\left(\mathbf{F}_{pj}\right) \right)}, \text{if } {i} \in \text{Top}^k_j(\mathbf{F})\label{eq:W}
	\end{equation}
	and 0 otherwise. $\text{Top}^k_j(\mathbf{F})$ denotes the set of indices that are top-$k$ in the $j$-th column of $\mathbf{F}$.
	These attentional weights are used to compute a weighted-sum of $\mathbf{v}^M$. For query position $j$, the feature $\mathbf{m}_j$ is read from memory by:
	\vspace{-0.1in}
	\begin{equation}
		\mathbf{m}_j = \sum_{p}^{THW} \mathbf{v}^M_p \mathbf{W}_{pj}
		\vspace{-0.1in}
	\end{equation}
	The read features will be concatenated with $\mathbf{v}^Q$ and passed to the decoder to generate the object mask. Skip-connections (not shown for clarity) from the query encoder to the decoder help to create a more accurate mask. 
	The output of the decoder is a stride 4 mask which is bilinearly upsampled to the original resolution. When there are multiple objects, we process each object one by one and combine the masks using soft aggregation~\cite{oh2019videoSTM}.
	\vspace{-0.05in}	
	\paragraph{Propagation strategy}
	\vspace{-0.05in}	
	Figure~\ref{fig:propagation} illustrates our bidirectional propagation strategy, similar to~\cite{oh2019fastInteractive}. Given a user-interacted reference frame $M^r_{t^r}$, we bidirectionally propagate the segmentation to other frames with two (forward and backward) independent passes.
	Given that each interacted frame is sufficiently well-annotated (which is more easily satisfied under our decoupled framework), the propagation stops once hitting a previously interacted frame or the end of the sequence. Following STM~\cite{oh2019videoSTM}, every 5th frame will be included and cached in the memory bank.
	The frame immediately before the query frame will also be included as temporary memory.
	In interactive settings, all user-interacted frames are trusted and added to the memory bank.
	
	\vspace{-4mm}
	\paragraph{Evaluation}
	The propagation module can be isolated for evaluation in a semi-supervised VOS setting (where the first-frame ground-truth segmentation is propagated to the entire video). 
	Table~\ref{tab:prop_ablation} tabulates our validation of the effectiveness of top-$k$ filtering (our new dataset \textbf{BL30K} to be detailed in Section~\ref{dataset}).
	The algorithm is not particularly sensitive to the choice of $k$ with similar performance for $k =20$ through $100$. $k=50$ in all our experiments. In principle, the value of $k$ should be linear to the image resolution such that the effective area after filtering is approximately the same. With top-$k$ filtering, our multi-object propagation runs at 11.2 FPS on a 2080Ti.
	
	\begin{table}[h]
		\vspace{-0.10in}
		\centering
		\begin{tabular}{l|c|c|c@{}l}
			\hline
			Methods & Top-$k$? & BL30K? & \multicolumn{2}{c}{$\mathcal{J}\&\mathcal{F}$} \\
			\Xhline{3\arrayrulewidth}
			RGMP~\cite{oh2018fastRGMP} & - & - & \quad$66.7$ &\\
			FEELVOS~\cite{voigtlaender2019feelvos} & - & - & \quad$71.5$ &\\
			PReMVOS~\cite{luiten2018premvos} & - & - & \quad$77.8$ &\\
			STM~\cite{oh2019videoSTM} & - & - & \quad$81.8$ &\\
			CFBI~\cite{yang2020collaborativeCFBI} & - & - & \quad$81.9$ &\\
			KMN~\cite{seong2020kernelizedMemory} & - & - & \quad$82.8$ &\\
			GraphMem~\cite{lu2020videoGraphMem} & - & - & \quad$82.8$ &\\
			\hline
			Ours & \xmark & \xmark & \quad$81.5$ & $_{-}$ \\ 
			Ours & \xmark & \cmark & \quad$83.8$ & $_{\uparrow2.3}$ \\ 
			Ours & \cmark & \xmark & \quad$83.1$ & $_{\uparrow1.6}$ \\ 
			Ours & \cmark & \cmark & \quad$\mathbf{84.5}$ & $_\mathbf{{\uparrow3.0}}$ \\ 
			\hline
		\end{tabular}
		\caption{Evaluation of our propagation module in the DAVIS 2017 multi-object semi-supervised validation set. Both top-$k$ filtering and BL30K are effective in increasing the performance. 
		$\uparrow$ indicates improvement over our baseline. 
		In addition, we obtain $76.5~\mathcal{J}\&\mathcal{F}$ on the DAVIS {\tt\small test-dev} set which is more difficult with harsh lighting conditions. Refer to the project website for more results.
		}
		\label{tab:prop_ablation}
		\vspace{-0.15in}
	\end{table}
	
	\subsection{Difference-Aware Fusion}\label{fusion}
	If the propagation ends with hitting a previously interacted frame $t^c$, there may exist conflicts in frames within $t^c$ and $t^r$. Fusion is thus required between the current propagated mask $M^{r'}$ and the previous mask results $M^{r-1}$.
	Previous approaches~\cite{oh2019fastInteractive,Yuk2020IVOSGlobalLocal} often employ a linear weighting scheme which is agnostic to the correction made and thus fails to capture the user's intent. 
	Oftentimes, the user correction will disappear mid-way between  $t^r$ and $t^c$.
	
\begin{figure}[t]
		\begin{center}
			\includegraphics[width=\linewidth]{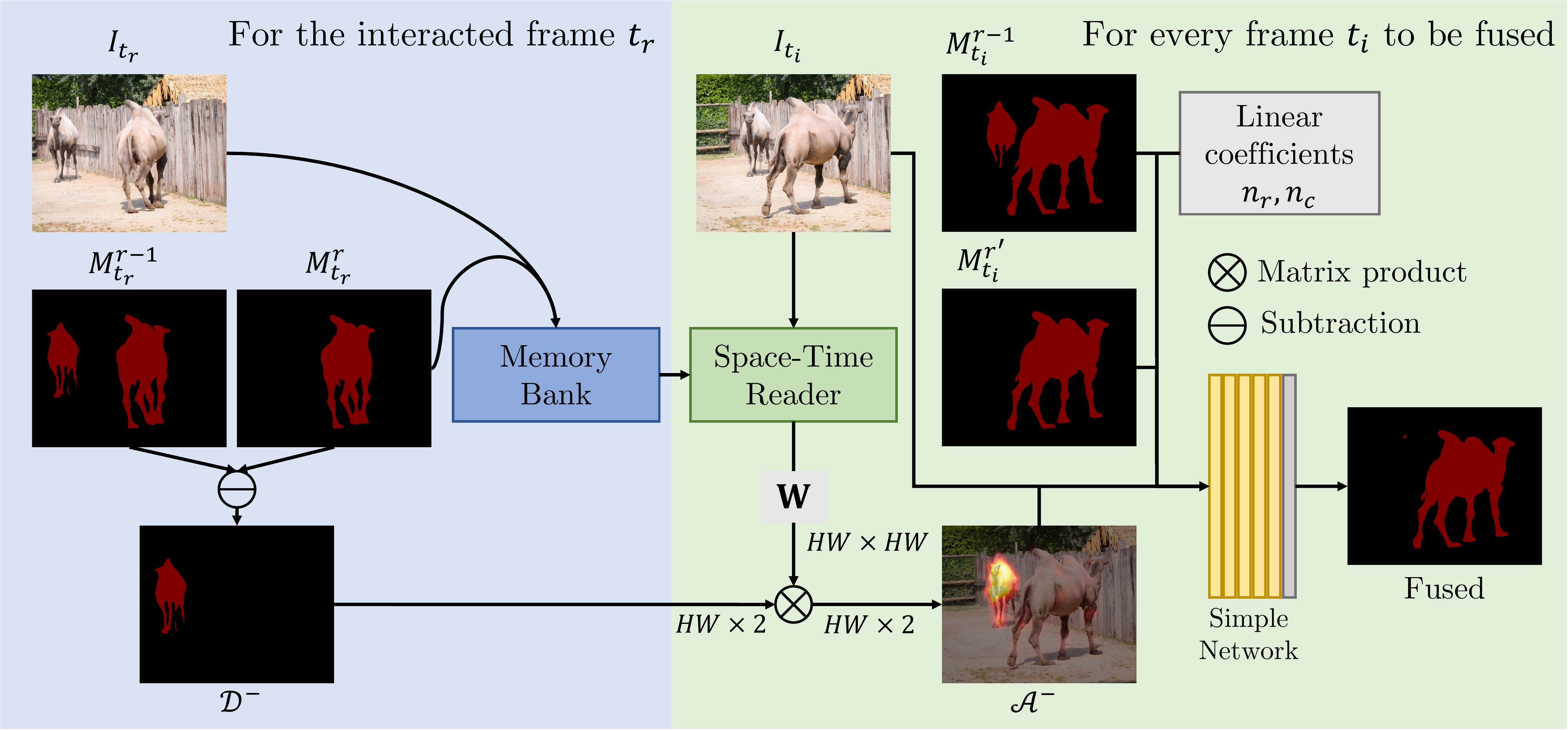}
		\end{center}
		\vspace{-0.15in}
		\caption{
			Mechanism of the difference-aware fusion module. The current propagated mask $M_{t_i}^{r'}$ at frame $I_{t_i}$ is fused with the previous mask $M^{r-1}_{t_i}$, guided by the mask difference from interaction at frame $t_r$. Only the negative terms $\mathcal{D}^-, \mathcal{A}^-$ are shown here for clarity. Note that although a correct mask is captured in $M_{t_i}^{r'}$, it is non-trivial to pick it up in the fusion step as shown in Figure~\ref{fig:fusion_result}.
		}
		\label{fig:fusion}
		\vspace{-0.15in}
	\end{figure}

	As illustrated in Figure~\ref{fig:fusion}, we propose a novel learnable fusion module that can keep the user correction in mind during fusion. Specifically, the user correction is captured as the differences in the mask before and after the user interaction at frame $t^r$:
	\begin{equation}
		\begin{aligned}[c]
			\mathcal{D}^+ = \left(M^{r}_{t^r} - M^{r-1}_{t^r}\right)_+
		\end{aligned}
		\quad
		\begin{aligned}[c]
			\mathcal{D}^- = \left(M^{r-1}_{t^r} - M^{r}_{t^r}\right)_+
		\end{aligned}
	\end{equation}
	where $(\cdot)_+$ is the $\max(\cdot, 0)$ operator. We compute the positive and negative changes separately as two masks $\mathcal{D}^+$ and $\mathcal{D}^-$. To fuse $t_i$, which is between $t^r$ and $t^c$, these masks cannot be used directly as they are not aligned with the target frame $t_i$. 
	The key insight is that we can leverage the affinity matrix $\mathbf{W}$ in Eq.~(\ref{eq:W}) computed by our space-time memory reader (Figure~\ref{fig:spacetime}) for correspondence matching. The interacted frame $t^r$ and target frame $t_i$ are used as memory and query respectively. 
	The aligned masks are computed by two matrix products:
	\begin{equation}
		\begin{aligned}[c]
			\mathcal{A}^+ = \mathbf{W}\mathcal{D}^+
		\end{aligned}
		\quad
		\begin{aligned}[c]
			\mathcal{A}^- = \mathbf{W}\mathcal{D}^-
		\end{aligned}
	\end{equation}
	Where $\mathcal{D}^+$ and $\mathcal{D}^-$ are downsampled using area averaging to match the image stride of $\mathbf{W}$, and the results are upsampled bilinearly to the original resolution.
	Additionally, traditional linear coefficients are also used to model possible decay during propagation:
	\begin{equation}
		\begin{aligned}[c]
			n_r = \frac{|t_i-t_r|}{|t_c-t_r|}
		\end{aligned}
		\quad
		\begin{aligned}[c]
			n_c = \frac{|t_i-t_c|}{|t_c-t_r|}
		\end{aligned}
	\end{equation}
	Note that $n_r+n_c=1$.
	Finally, the set of features $(I_{t_i},  M_{t_i}^{r'}, M_{t_i}^{r-1}, \mathcal{A}^+, \mathcal{A}^-, n_r, n_c )$ are fed into a simple five-layer residual network which is terminated by a sigmoid to output a final fused mask. 
	
\begin{figure}[t]
		\begin{center}
			\small
			\begin{tabular}{@{\hspace{0mm}}c@{\hspace{0.5mm}}c@{\hspace{0.5mm}}c@{\hspace{0.5mm}}c}
				\includegraphics[width=.24\linewidth]{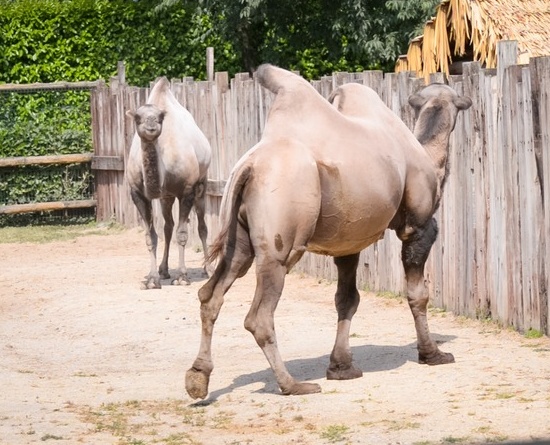} &
				\includegraphics[width=.24\linewidth]{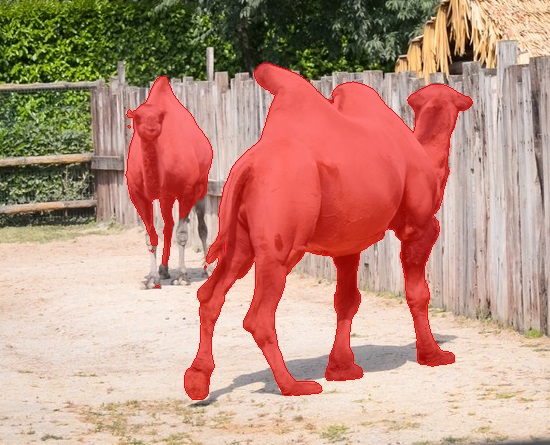} &
				\includegraphics[width=.24\linewidth]{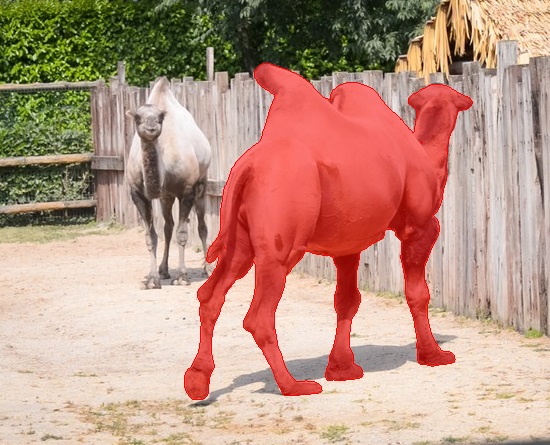} &
				\includegraphics[width=.24\linewidth]{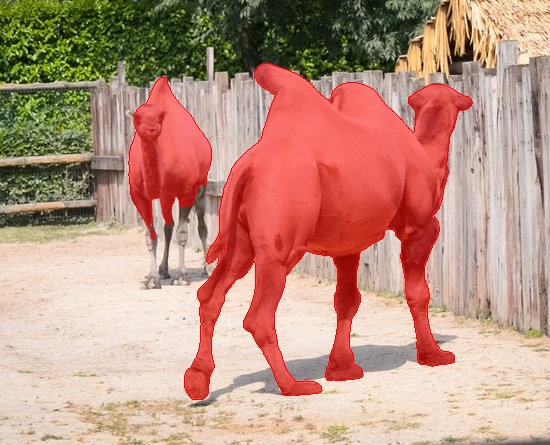} \\
				(a) $I_{t_i}$ &
				(b) $M^{r-1}_{t_i}$ &
				(c) $M^{r'}_{t_i}$ &
				(d) Linear \\
				\includegraphics[width=.24\linewidth]{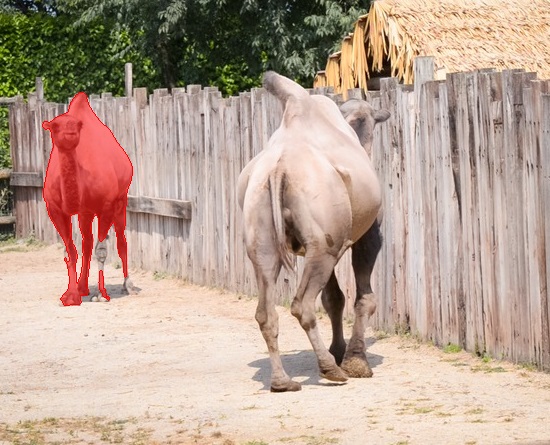} &
				\includegraphics[width=.24\linewidth]{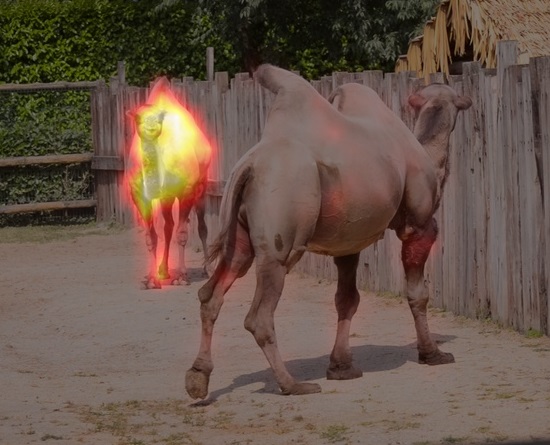} &
				\includegraphics[width=.24\linewidth]{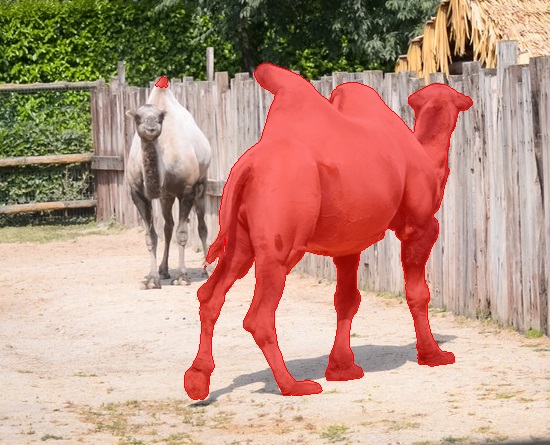} &
				\includegraphics[width=.24\linewidth]{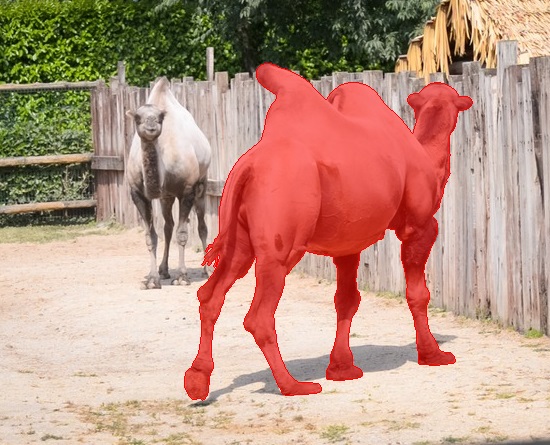}
				\\
				(e) $\mathcal{D}^-$ &
				(f) $\mathcal{A}^-$ &
				(g) Fused &
				(h) GT
			\end{tabular}
		\end{center}
		\vspace{-0.10in}
		\caption{
			Continuing Figure~\ref{fig:fusion}, showing popularly used linear blending is insufficient. Suppose the user first annotates $t_c=25$, then corrects the mask at $t_r=89$. For the query frame with $t_i=51$ which is closer to $25$ than to $89$, linear blending (or any symmetric function that only uses the temporal distance) fails in (d). With our difference aware fusion, we use the mask difference (e) to form an aligned attention (f) that captures the correction. Our result is shown in (g).
		}
		\label{fig:fusion_result}
		\vspace{-0.15in}
	\end{figure}

	As illustrated in Figure~\ref{fig:fusion_result}, our fusion method can capture the user's intention as an aligned attention map, which allows our algorithm to propagate corrections beyond the mid-point. Such fusion cannot be achieved in previous linear-blending methods~\cite{oh2019fastInteractive,Yuk2020IVOSGlobalLocal} (non-symmetric blending~\cite{Yuk2020IVOSGlobalLocal} will fail if we swap the order of interaction).
	Evaluation of the fusion module is presented in Section~\ref{ablation}.
	
	\vspace{-2mm}
	\section{Dataset: BL30K}\label{dataset}
	High-quality VOS datasets are expensive to collect at a large scale -- DAVIS~\cite{Caelles_arXiv_2019} is high-quality yet lacks quantity; YouTubeVOS~\cite{xu2018youtubeVOS} is large but has moderate quality annotations. In this paper we contribute a new synthetic VOS dataset \textbf{BL30K} that not only is large-scale but also provides pixel-accurate segmentations. Table~\ref{tab:multiclass_table} compares the three datasets. 
	
	\begin{table}[h]
		\small
		\centering
		\vspace{-0.07in}
		\begin{tabular}{l|c|c|c}
			\hline
			Dataset & \# Videos & \# Frames & Label Quality \\
			\Xhline{3\arrayrulewidth}
			DAVIS~\cite{Caelles_arXiv_2019} & 90 & 6,208 & High \\
			\hline
			YV~\cite{xu2018youtubeVOS} & 3,471 & 94,588 & Moderate \\
			\hline
			{\bf BL30K} & 29,989 & 4,783,680 & High \\
			\hline
		\end{tabular}
		\caption{Comparison between different VOS datasets. Only frames with publicly available ground truths are counted.}
		\label{tab:multiclass_table}
		\vspace{-0.15in}
	\end{table}
	
\begin{figure}[h]
		\vspace{-0.07in}
		\begin{center}
			\includegraphics[width=0.24\linewidth]{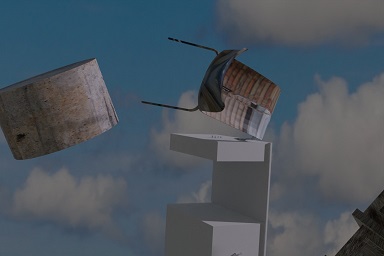}
			\includegraphics[width=0.24\linewidth]{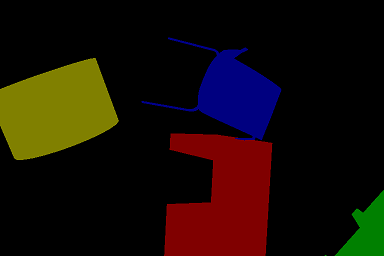}
			\includegraphics[width=0.24\linewidth]{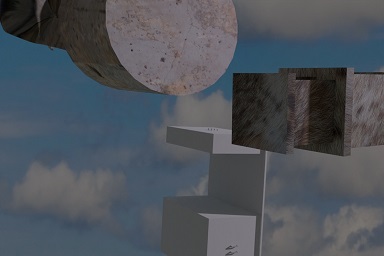}
			\includegraphics[width=0.24\linewidth]{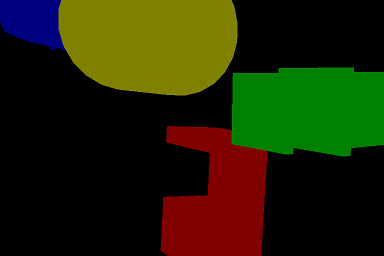}
		\end{center}
		\vspace{-0.15in}
		\caption{Sample data from the BL30K dataset.}
		\label{fig:bldata}
		\vspace{-0.22in}
	\end{figure}

	Using an open-source rendering engine \textit{Blender}~\cite{Blender, denninger2019blenderproc}, we animate 51,300 three-dimensional models from ShapeNet~\cite{shapenet2015} and produce the corresponding RGB images and segmentations with a two-pass rendering scheme. Background images and object textures are collected using Google image search to enrich the dataset. Each video consists of 160 frames with a resolution of $768\times512$. Compared with FlythingThings3D~\cite{mayer2016largeFlyingThings3D}, our videos have a higher frame rate and a much longer sequence length, making ours suitable for the VOS task while~\cite{mayer2016largeFlyingThings3D} is not applicable.
	Figure~\ref{fig:bldata} shows one sample in our dataset.
	To the best of our knowledge, BL30K is the largest publicly available VOS dataset to date. Despite that the dataset being synthetic, it does significantly help in improving real-world performance as shown in our ablation study (Section~\ref{ablation}). Note that this gain is \emph{not} simply caused by more training iterations as extended training on YouTubeVOS~\cite{xu2018youtubeVOS} and DAVIS~\cite{Caelles_arXiv_2018} leads to severe overfitting in our experiments.
	
	\section{Implementation Details}
	All three modules can be efficiently trained using just two 11GB GPU with the Adam optimizer~\cite{kingma2015adam}. The propagation module is first trained on synthetic video sequences from static images 
	following~\cite{oh2019videoSTM}, which is then transferred to BL30K, YouTubeVOS~\cite{xu2018youtubeVOS} and DAVIS~\cite{Caelles_arXiv_2018}. In each training iteration, we pick three random frames in a video sequence, with the maximum distance between frames increased from 5 to 25 gradually (curriculum learning) and annealed back to 5 toward the end of training~\cite{zhangspatialWorkshop}. 
	The S2M module is independently trained on static images only. 
	The fusion module is trained with the output of a pretrained propagation module, first on BL30K, and then transferred to DAVIS~\cite{Caelles_arXiv_2018}. YouTubeVOS~\cite{xu2018youtubeVOS} is not used here due to its less accurate annotation. Table~\ref{tab:run_time} tabulates the running time of different components in our model. Refer to our open-sourced code for detailed hyperparameters settings.
	It takes about two weeks to train all the modules with two GPUs.
	
	\begin{table}[h]	
		\begin{center}
			\small
			\begin{tabular}{l|c}
				\hline
				& Time (ms) / frame / instance\\
				\Xhline{3\arrayrulewidth}
				Scribble-to-Mask (S2M) & 29 \\
				\hline
				f-BRS~\cite{sofiiuk2020fbrs} & $\sim$60 \\
				\hline
				Propagation w/o top-$k$ & 51 \\
				\hline
				Propagation w/ top-$k$ & 44 \\
				\hline
				Fusion & 9 \\
				\hline
			\end{tabular}
		\end{center}
		\vspace{-0.15in}
		\caption{Running time analysis of each component in our model. Time is measured on the 480p DAVIS 2017 validation set; time for propagation is amortized. 
		For an average of two objects in DAVIS 2017, our baseline performance matches the one reported in STM~\cite{oh2020STMPAMI}. 
		Run time of f-BRS depends on the input as adaptive optimization is involved.
		Note that propagation is performed sparsely which keep our algorithm the fastest among competitors.}
		\label{tab:run_time}
	\end{table}
	
	\vspace{-2mm}
	\section{Experiments}\label{expr}
	
	\subsection{DAVIS Interactive Track}
	In the DAVIS 2020 Challenge~\cite{Caelles_arXiv_2019} interactive track, the robot first provides scribbles for a selected frame, waits for the algorithm's output, and then provides corrective scribbles for the worst frame of all the candidate frames listed by the algorithm. The above is repeated up to 8 rounds. 
	To demonstrate the effectiveness of our proposed decoupled method which requires less temporally dense interactions, we limit ourselves to interact with only three frames. 
	Specifically, we force the robot to only pick a new frame in the \nth{1}, \nth{4}, and \nth{7} interactions. Our algorithm stays in an instant feedback loop for the same frame and performs propagation only when the robot has finished annotating one frame. Note that this behavior can be implemented without altering the official API.
	
\begin{figure}[h]
		\begin{center}
			\includegraphics[width=0.99\linewidth]{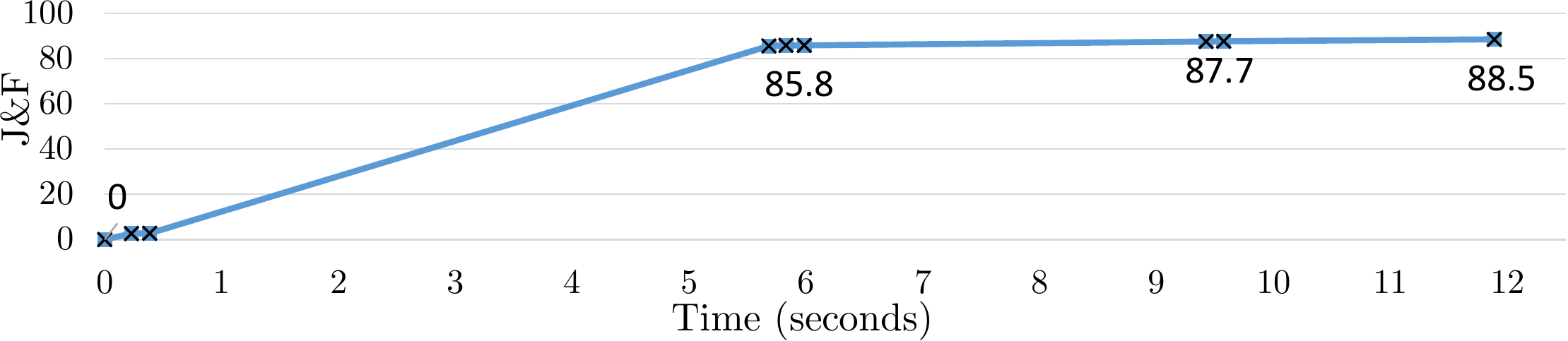}
		\end{center}
		\vspace{-0.15in}
		\caption{$\mathcal{J}$\&$\mathcal{F}$ performance on the DAVIS validation set. Clustered points represent real-time corrections in the instant feedback loop; each cluster represents a frame switch and propagation. Our method is highly efficient, achieving better performance in $\sim$12~seconds on average compared with 55+ seconds in~\cite{Yuk2020IVOSGlobalLocal} or 37~seconds in~\cite{oh2020STMPAMI}.}
		\label{fig:jf_vs_round}
		\vspace{-0.15in}
	\end{figure}

\begin{figure*}[t]
		\vspace{-0.25in}
		\begin{center}
			\begin{tabular}{m{1em}@{\hspace{0mm}}c@{\hspace{-.5mm}}c@{\hspace{-.5mm}}c@{\hspace{-.5mm}}c@{\hspace{-.5mm}}c}
				\rotatebox[origin=c]{90}{ATNet}&
				\raisebox{-0.5\height}{
					\begin{overpic}[width=0.19\linewidth,percent]{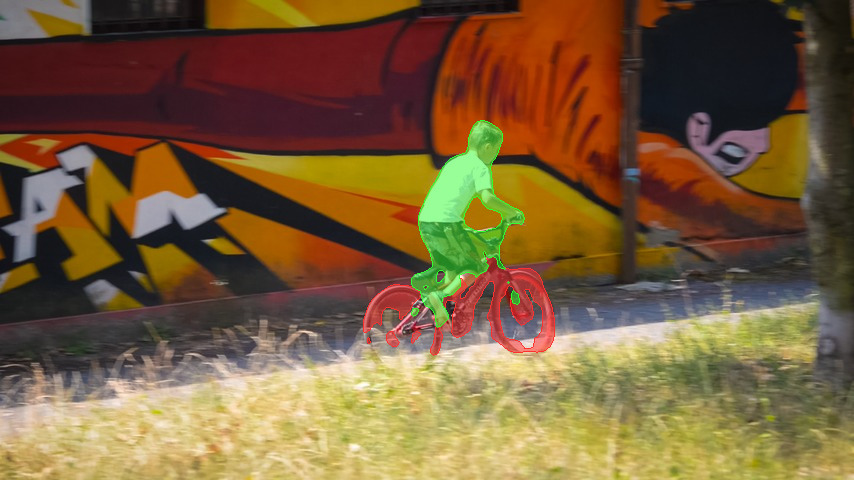}
					\end{overpic}
				}&
				\raisebox{-0.5\height}{
					\begin{overpic}[width=0.19\linewidth,percent]{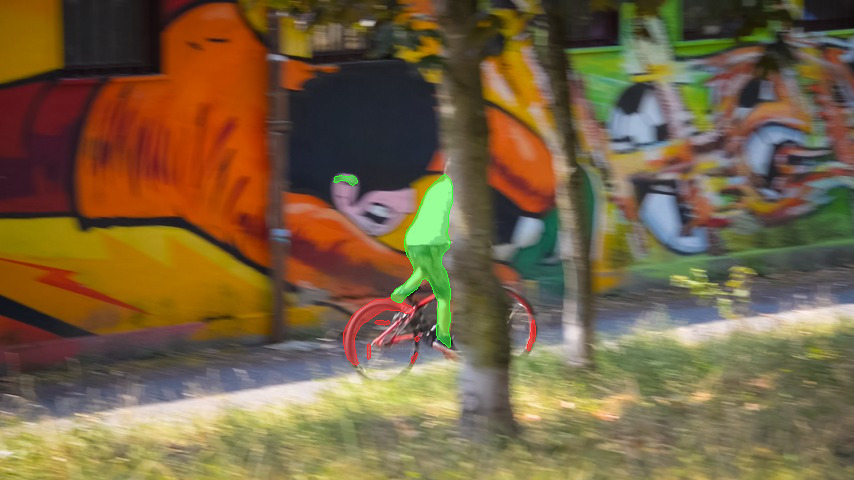}
						\put(0,36){\color{yellow}\fbox{\includegraphics[scale=0.21]{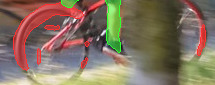}}}
						\put(38,12){\color{yellow}\fbox{\makebox(25,10){}}}
					\end{overpic}
				}&
				\raisebox{-0.5\height}{
					\begin{overpic}[width=0.19\linewidth,percent]{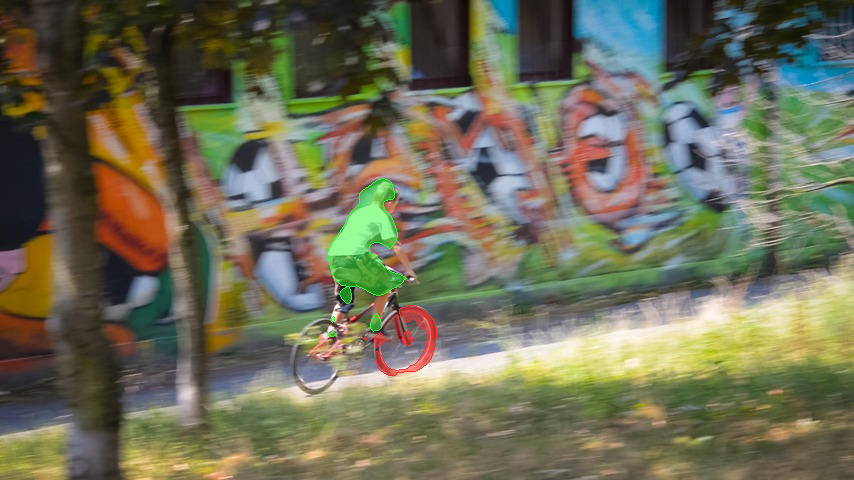}
						\put(57,33){\color{yellow}\fbox{\includegraphics[scale=0.3]{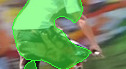}}}
						\put(33,21){\color{yellow}\fbox{\framebox(17,9){}}}
					\end{overpic}
				}&
				\raisebox{-0.5\height}{
					\begin{overpic}[width=0.19\linewidth,percent]{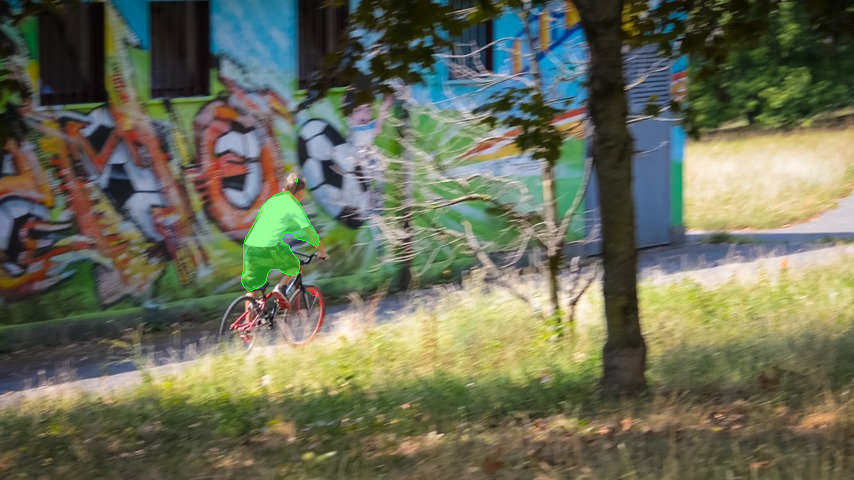}
						\put(62,4){\color{yellow}\fbox{\includegraphics[scale=0.28]{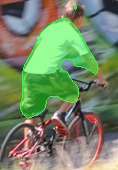}}}
						\put(25,15){\color{yellow}\fbox{\framebox(13,21){}}}
					\end{overpic}
				}&
				\raisebox{-0.5\height}{
					\begin{overpic}[width=0.19\linewidth,percent]{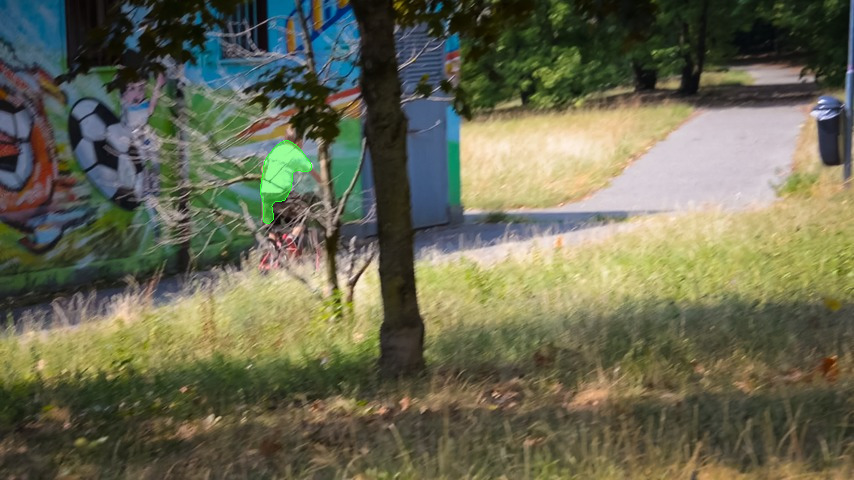}
						\put(72,2){\color{yellow}\fbox{\includegraphics[scale=0.32]{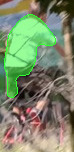}}}
						\put(28,23){\color{yellow}\fbox{\framebox(10,18){}}}
					\end{overpic}
				}\\
				\rotatebox[origin=c]{90}{Ours}&
				\raisebox{-0.5\height}{\includegraphics[width=0.19\linewidth]{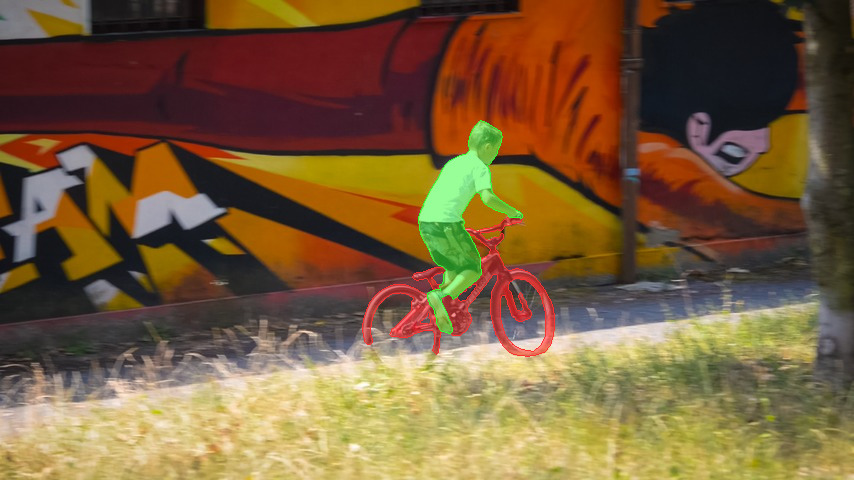}}&
				\raisebox{-0.5\height}{
					\begin{overpic}[width=0.19\linewidth,percent]{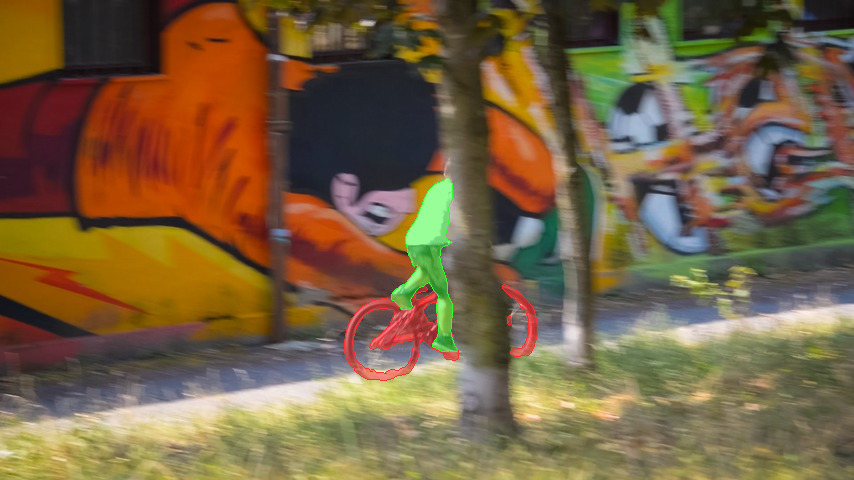}
						\put(0,37){\color{yellow}\fbox{\includegraphics[scale=0.21]{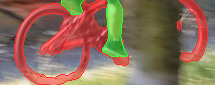}}}
						\put(38,12){\color{yellow}\fbox{\framebox(25,10){}}}
					\end{overpic}
				}&
				\raisebox{-0.5\height}{
					\begin{overpic}[width=0.19\linewidth,percent]{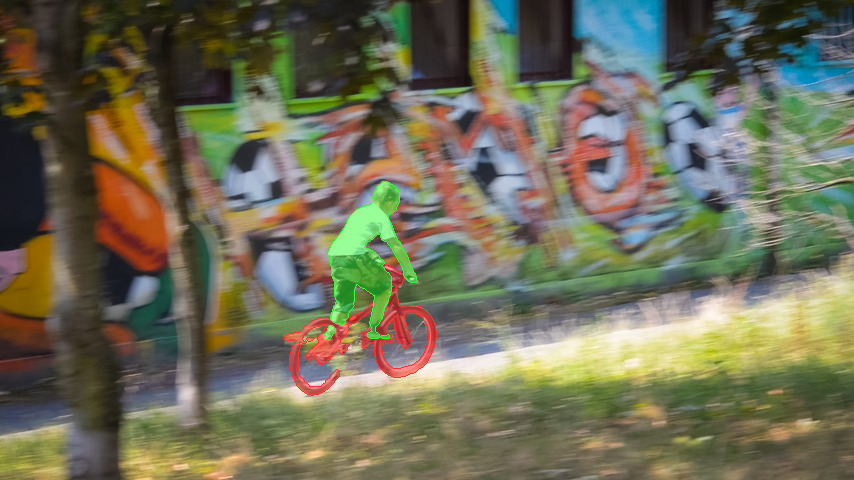}
						\put(57,33){\color{yellow}\fbox{\includegraphics[scale=0.3]{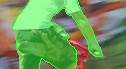}}}
						\put(33,21){\color{yellow}\fbox{\framebox(17,9){}}}
					\end{overpic}
				}&
				\raisebox{-0.5\height}{
					\begin{overpic}[width=0.19\linewidth,percent]{results/bmx/ours/00045.jpg}
						\put(62,4){\color{yellow}\fbox{\includegraphics[scale=0.28]{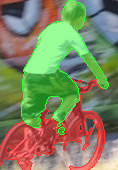}}}
						\put(25,15){\color{yellow}\fbox{\framebox(13,21){}}}
					\end{overpic}
				}&
				\raisebox{-0.5\height}{
					\begin{overpic}[width=0.19\linewidth,percent]{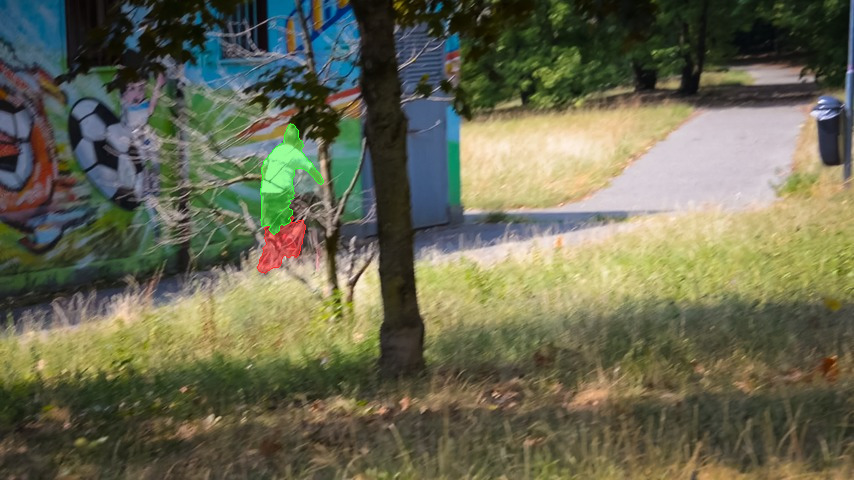}
						\put(72,2){\color{yellow}\fbox{\includegraphics[scale=0.32]{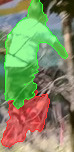}}}
						\put(28,23){\color{yellow}\fbox{\framebox(10,18){}}}
					\end{overpic}
				}\\
				\rotatebox[origin=c]{90}{ATNet}&
				\raisebox{-0.5\height}{\includegraphics[width=0.19\linewidth]{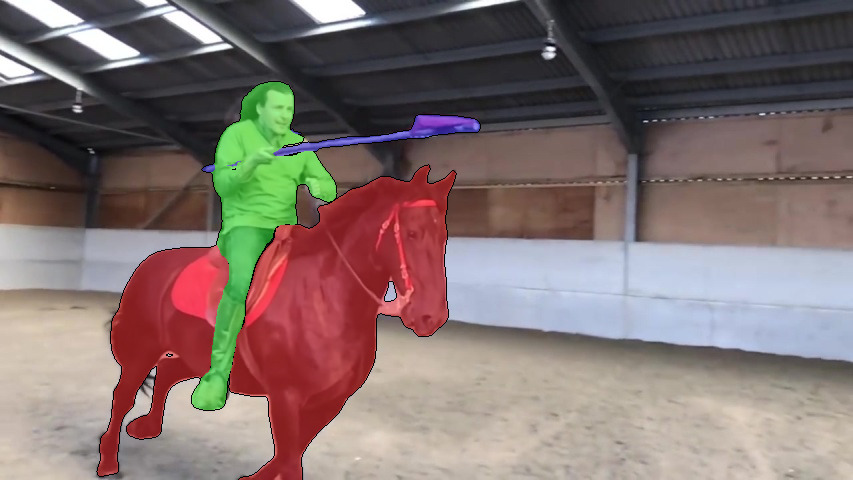}}&
				\raisebox{-0.5\height}{
					\begin{overpic}[width=0.19\linewidth,percent]{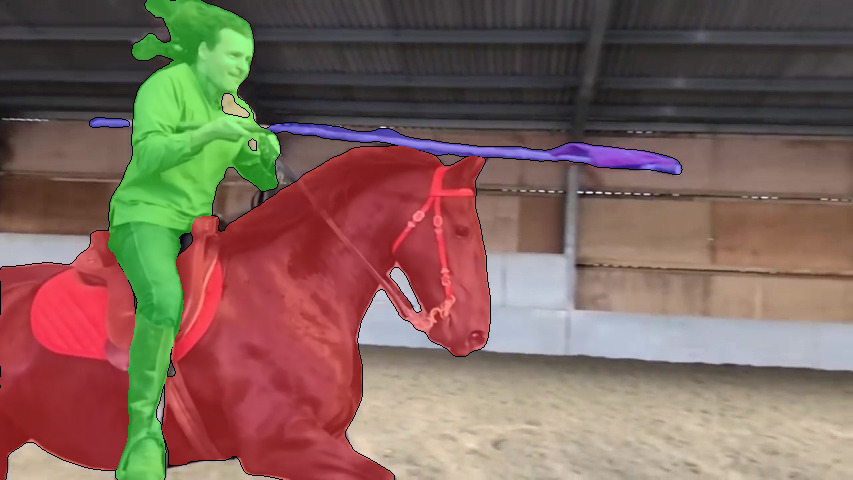}
						\put(59,17){\color{yellow}\fbox{\includegraphics[scale=0.30]{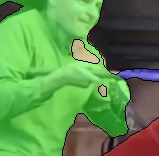}}}
						\put(16,32){\color{yellow}\fbox{\framebox(17,17){}}}
					\end{overpic}
				}&
				\raisebox{-0.5\height}{
					\begin{overpic}[width=0.19\linewidth,percent]{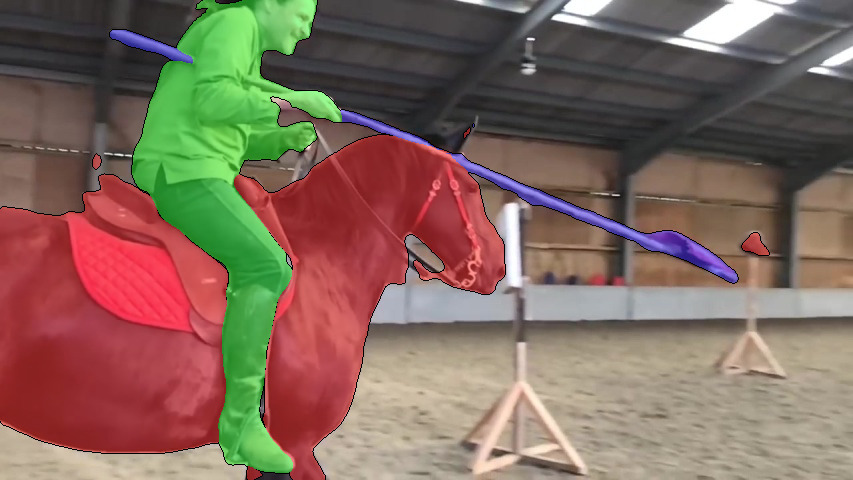}
						\put(63,25){\color{yellow}\fbox{\includegraphics[scale=0.35]{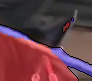}}}
						\put(47,35){\color{yellow}\fbox{\framebox(10,9){}}}
					\end{overpic}
				}&
				\raisebox{-0.5\height}{
					\begin{overpic}[width=0.19\linewidth,percent]{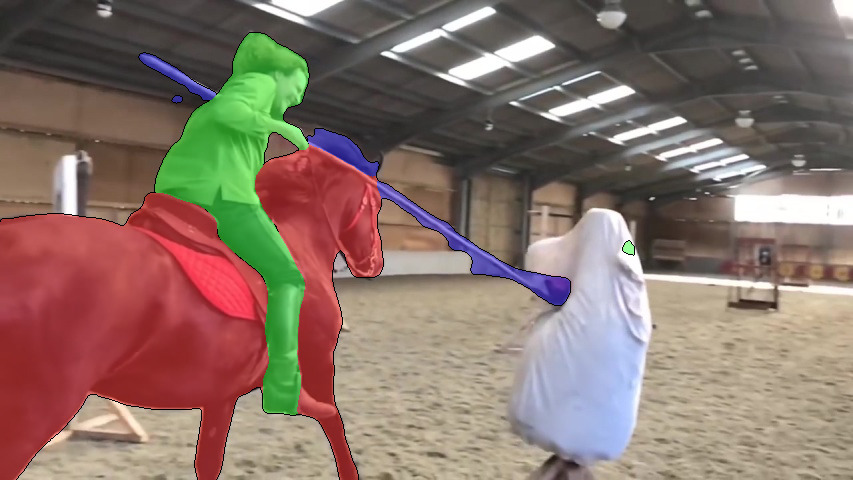}
						\put(67,25){\color{yellow}\fbox{\includegraphics[scale=0.23]{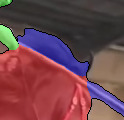}}}
						\put(34,31){\color{yellow}\fbox{\framebox(12,12){}}}
					\end{overpic}
				}&
				\raisebox{-0.5\height}{
					\begin{overpic}[width=0.19\linewidth,percent]{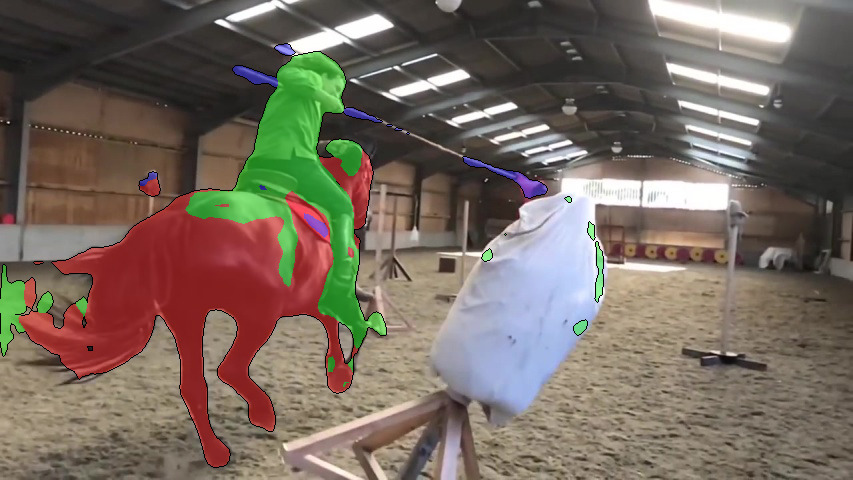}
						\put(64,35){\color{yellow}\fbox{\includegraphics[scale=0.23]{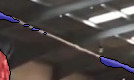}}}
						\put(42,35){\color{yellow}\fbox{\framebox(14,8){}}}
					\end{overpic}
				}\\
				\rotatebox[origin=c]{90}{Ours}&
				\raisebox{-0.5\height}{\includegraphics[width=0.19\linewidth]{results/lance/atnet/00000.jpg}}&
				\raisebox{-0.5\height}{
					\begin{overpic}[width=0.19\linewidth,percent]{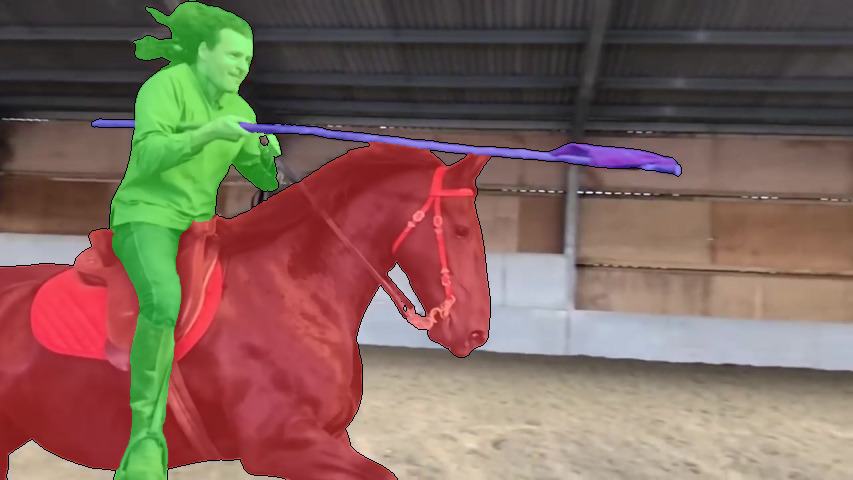}
						\put(59,17){\color{yellow}\fbox{\includegraphics[scale=0.30]{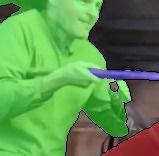}}}
						\put(16,32){\color{yellow}\fbox{\framebox(17,17){}}}
					\end{overpic}
				}&
				\raisebox{-0.5\height}{
					\begin{overpic}[width=0.19\linewidth,percent]{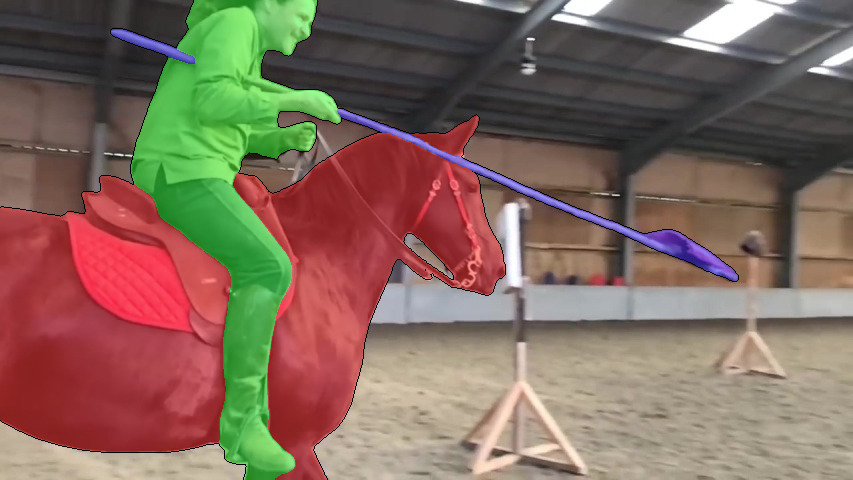}
						\put(63,25){\color{yellow}\fbox{\includegraphics[scale=0.35]{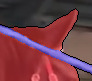}}}
						\put(47,35){\color{yellow}\fbox{\framebox(10,9){}}}
					\end{overpic}
				}&
				\raisebox{-0.5\height}{
					\begin{overpic}[width=0.19\linewidth,percent]{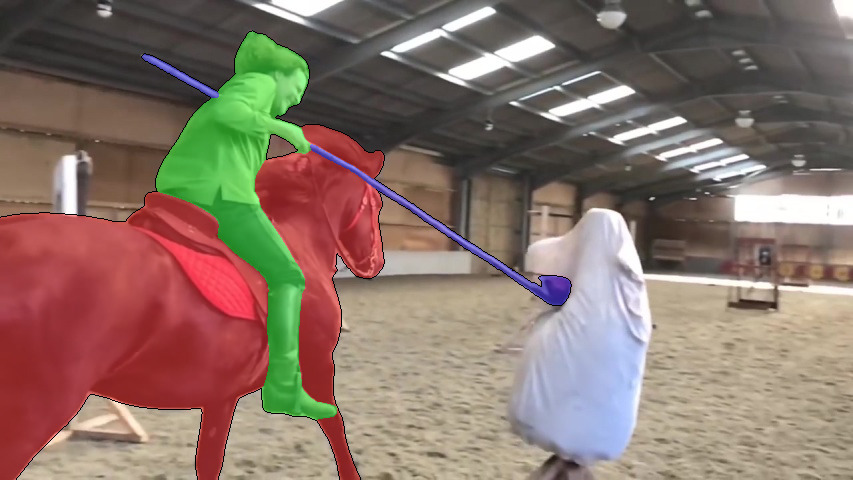}
						\put(67,25){\color{yellow}\fbox{\includegraphics[scale=0.23]{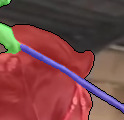}}}
						\put(34,31){\color{yellow}\fbox{\framebox(12,12){}}}
					\end{overpic}
				}&
				\raisebox{-0.5\height}{
					\begin{overpic}[width=0.19\linewidth,percent]{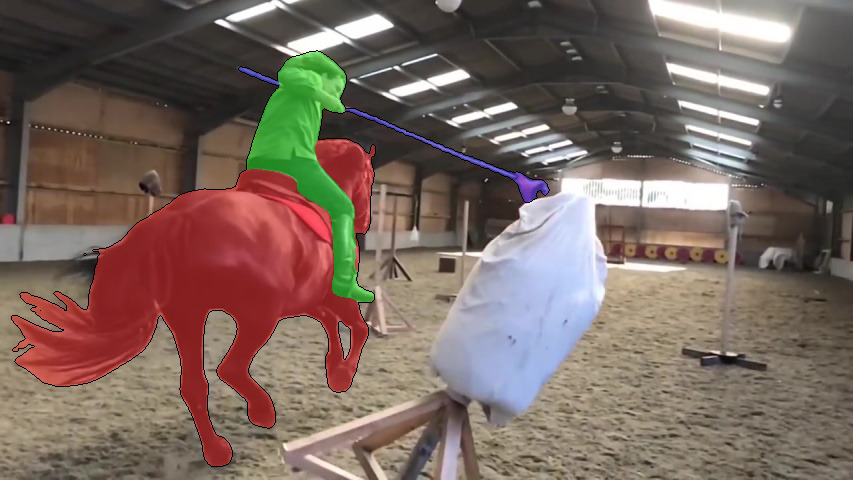}
						\put(64,35){\color{yellow}\fbox{\includegraphics[scale=0.23]{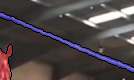}}}
						\put(42,35){\color{yellow}\fbox{\framebox(14,8){}}}
					\end{overpic}
				}\\
			\end{tabular}
			
			\vspace{0.5em}
			
			\begin{tabular}{c@{\hspace{1mm}}c@{\hspace{1mm}}c@{\hspace{1mm}}c@{\hspace{1mm}}c}
				\includegraphics[width=0.199\linewidth]{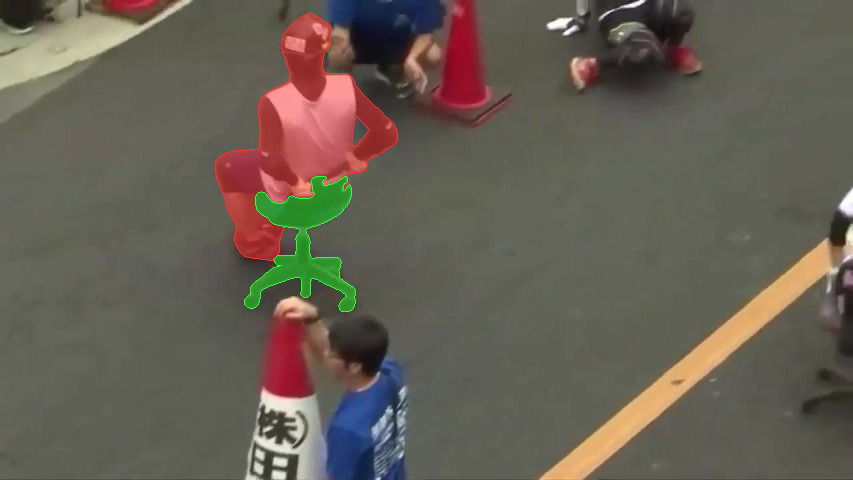}&
				\includegraphics[width=0.199\linewidth]{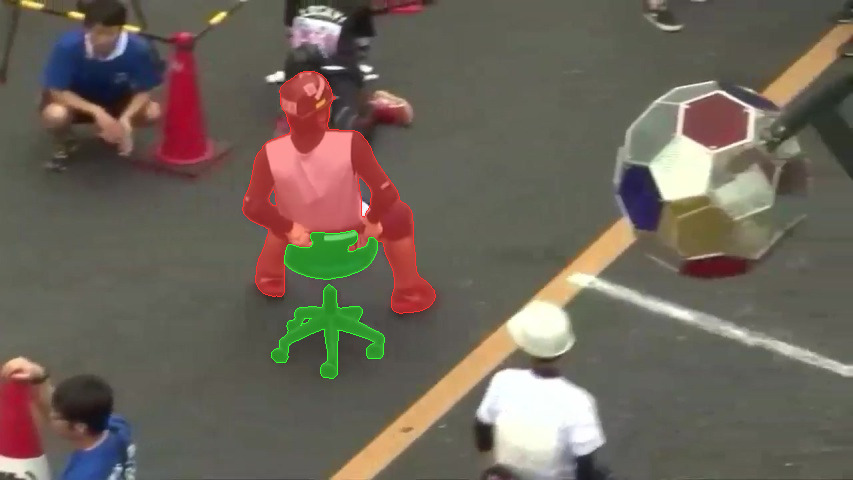}&
				\includegraphics[width=0.199\linewidth]{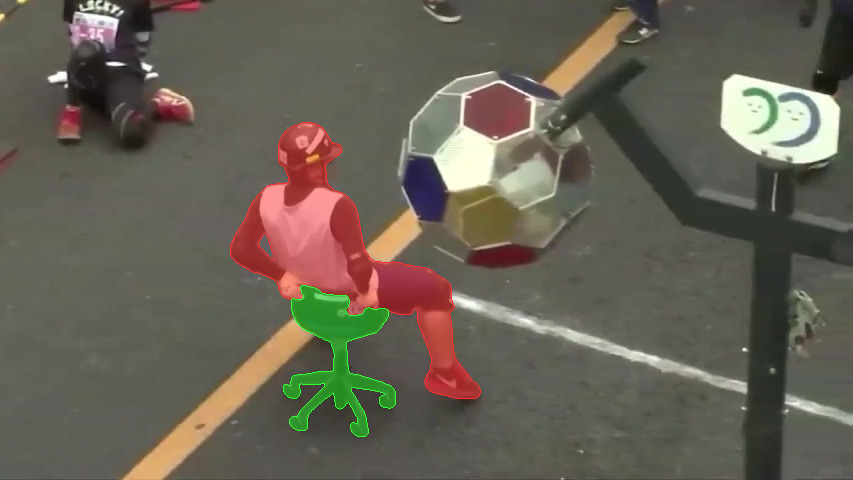}&
				\includegraphics[width=0.199\linewidth]{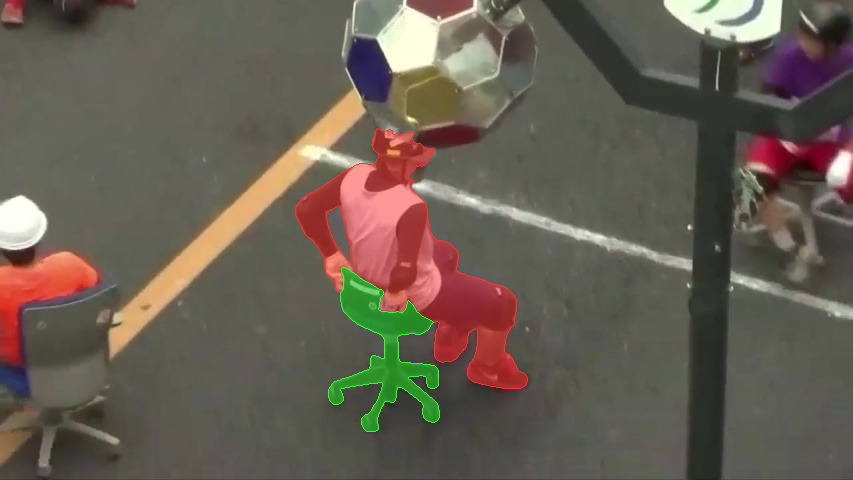}&
				\includegraphics[width=0.199\linewidth]{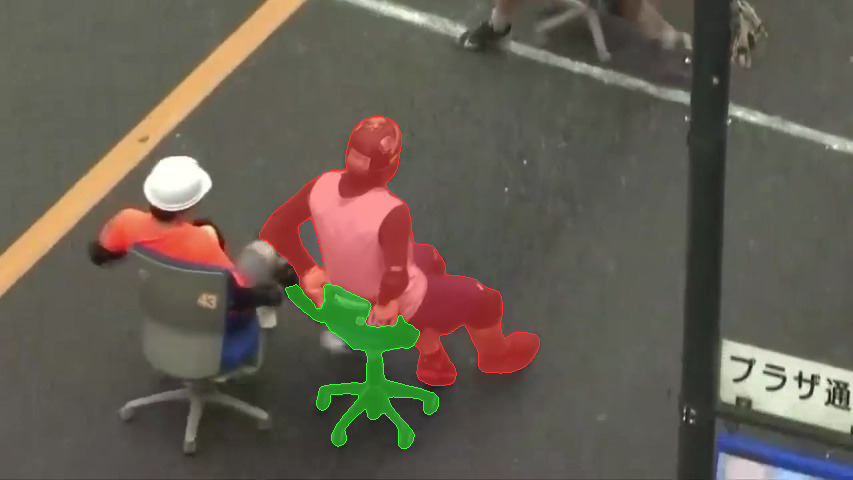}\\
				\includegraphics[width=0.199\linewidth]{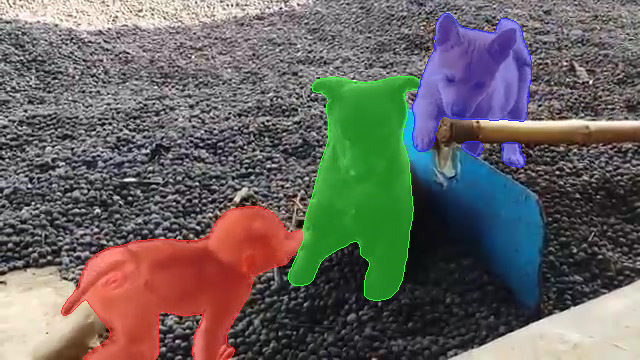}&
				\includegraphics[width=0.199\linewidth]{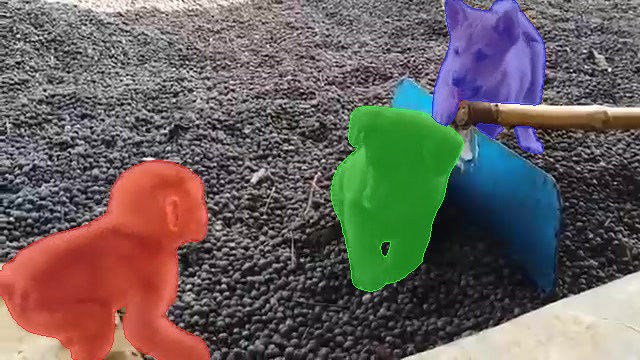}&
				\includegraphics[width=0.199\linewidth]{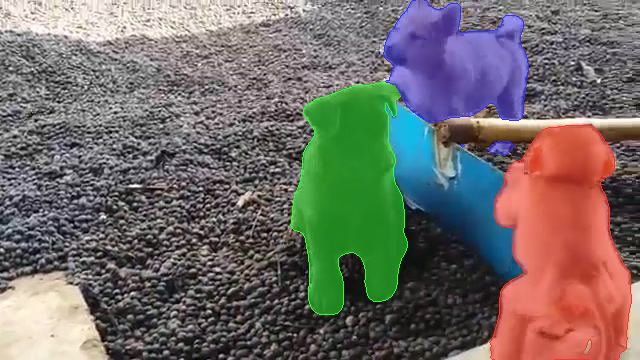}&
				\includegraphics[width=0.199\linewidth]{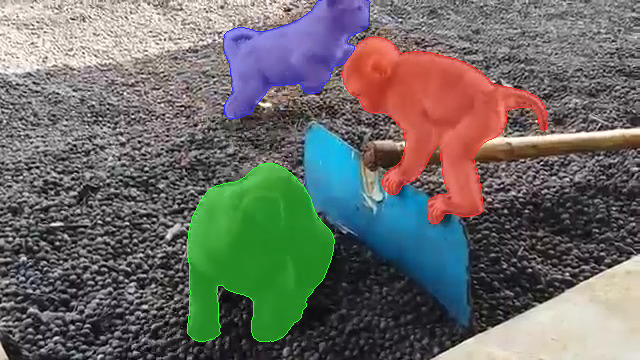}&
				\includegraphics[width=0.199\linewidth]{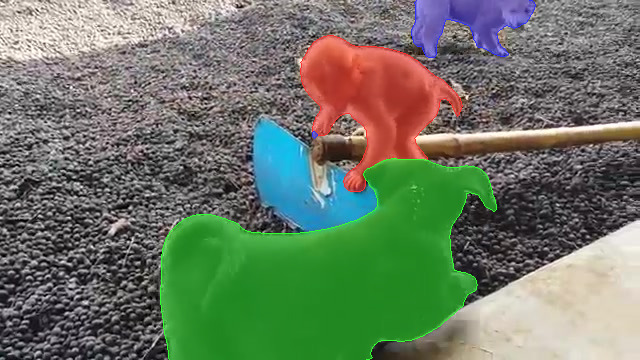}\\
			\end{tabular}
		\end{center}
		\vspace{-0.15in}
		\caption{Top four rows: Qualitative comparison of our method with ATNet~\cite{Yuk2020IVOSGlobalLocal} on the DAVIS interactive track (top two) and on previously unseen Internet video (middle two) with real user interactions (as detailed as possible on two frames). Bottom two rows: More results from our method on real-world videos from the Internet. Additional video results can be found on the project website. }
		\label{fig:visualization}
	\end{figure*}

	Table~\ref{tab:davis_interactive} tabulates the comparison results. 
	Figure~\ref{fig:jf_vs_round} plots the performance measured on $\mathcal{J}$\&$\mathcal{F}$ versus time. Note that, even with the above additional constraint, our method outperforms current state-of-the-art methods. We use the same GPU (RTX 2080Ti) as our closest competitor~\cite{oh2020STMPAMI}. Figure~\ref{fig:visualization} provides qualitative comparisons and visual results.
	
	\begin{table}[h]
		\vspace{-0.10in}
		\small
		\centering
		\begin{tabular}{l|c|c|c|c}
			\hline
			Methods & AUC-$\mathcal{J}$ & $\mathcal{J}^\text{\textdagger}$ & AUC-$\mathcal{J}$\&$\mathcal{F}$ & $\mathcal{J}$\&$\mathcal{F}^\text{\textdagger}$ \\
			\Xhline{3\arrayrulewidth}
			Oh \textit{et el.}~\cite{oh2019fastInteractive} & 69.1 & 73.4 & - & - \\
			\hline
			MANet~\cite{miao2020memoryAggregationInteractive} & 74.9 & 76.1 & - & - \\
			\hline
			ATNet~\cite{Yuk2020IVOSGlobalLocal} & 77.1 & 79.0 & 80.9 & 82.7 \\
			\hline
			STM~\cite{oh2020STMPAMI} & - & - & 80.3 & 84.8 \\
			\hline
			Ours & \textbf{84.9} & \textbf{85.4} & \textbf{87.9} & \textbf{88.5} \\
			\hline
		\end{tabular}
		\caption{Performance on the DAVIS interactive validation set. Our method outperforms all competitors while receiving only interactions in 3 frames instead of 8. \textdagger Interpolated value~@60s.}
		\label{tab:davis_interactive}
		\vspace{-0.15in}
	\end{table}
	
	\subsection{Ablation Study}\label{ablation}
	
	Table~\ref{tab:ablation} tabulates the quantitative evaluation on the effectiveness of BL30K and the fusion module. We show that 1)\ the proposed top-$k$ memory read transfers well to the interactive setting, 2)\ BL30K helps in real-world tasks despite being synthetic, and 3)\ Difference-aware fusion module outperforms na\"ive linear blending and difference-agnostic (learnable) fusion with the same network architecture. 
	Additionally, we show the upper bound performance of our method given perfect interaction masks. 
	
	\setlength{\fboxrule}{1pt}
	\setlength{\fboxsep}{0pt}

	\begin{table}[h]
	\vspace{-0.10in}
	\centering
	\begin{tabular}{l|c@{}l|c@{}l}
		\hline
		Model & \multicolumn{2}{c|}{AUC-$\mathcal{J}$\&$\mathcal{F}$} &  \multicolumn{2}{c}{$\mathcal{J}$\&$\mathcal{F}\text{@60s}$} \\
		\Xhline{3\arrayrulewidth}
		STM & \quad$80.3$ & & \quad$84.8$ & \\
		\Xhline{2\arrayrulewidth}
		Baseline & \quad$86.0$ & $_{-}$ & \quad$86.6$ & $_{-}$ \\
		\hline
		(+) Top-$k$ & \quad$87.2$ & $_{\uparrow 1.2}$ & \quad$87.8$ & $_{\uparrow 1.2}$ \\
		\hline
		(+) BL30K pretraining & \quad$87.4$ & $_{\uparrow 1.4}$ & \quad$88.0$ & $_{\uparrow 1.4}$ \\
		\hline
		(+) Learnable fusion  & \quad$87.6$ & $_{\uparrow 1.6}$ & \quad$88.2$ & $_{\uparrow 1.6}$ \\
		\hline
		(+) Difference-aware & \quad\multirow{2}{*}{$\mathbf{87.9}$} & \multirow{2}{*}{$\mathbf{_{\uparrow 1.9}}$} & \quad\multirow{ 2}{*}{$\mathbf{88.5}$} & \multirow{ 2}{*}{$\mathbf{_{\uparrow 1.9}}$} \\
		\quad\space\space (Full model) & & & & \\
		\Xhline{2\arrayrulewidth}
		Perfect interaction & \quad$90.2$ & & \quad$90.7$ & \\
		\hline
	\end{tabular}
	\caption{Ablation study on the DAVIS interactive validation set. 
		Our decoupled baseline already outperforms SOTA by a large margin. Despite the high baseline, we show that top-$k$ memory filtering, pretraining in the BL30K dataset, and the difference-aware fusion module can further improve its performance. In the last row, we replace the interaction module with an oracle that provides ground-truth masks to evaluate the upper-bound of our method given perfect interactions in 3 frames.}
	\label{tab:ablation}
	\vspace{-0.10in}
	\end{table}

	\subsection{User Study}\label{user_study}
	We conduct a user study to quantitatively evaluate user's preferences and  human effort required to label a video using iVOS algorithms. Specifically, we quantify the required human effort by the total {\em user time} which includes the time for interaction, searching, or pausing to think while excluding all computational time. 
	We linearly interpolate the IoU versus user-time graph and compute the area under curve~(AUC) for evaluation.
	We compare with ATNet~\cite{Yuk2020IVOSGlobalLocal} which is the best performing method with available source code to the best of our knowledge. We use two variants of our method -- one with S2M as the only interaction option~(Ours-S2M), and the other allows users to use a combination of S2M, f\nobreakdash-BRS~\cite{sofiiuk2020fbrs} and free-hand drawing, with the local control option~(Ours-Free).
	
\begin{figure}[h]
		\begin{center}
			\includegraphics[width=0.99\linewidth]{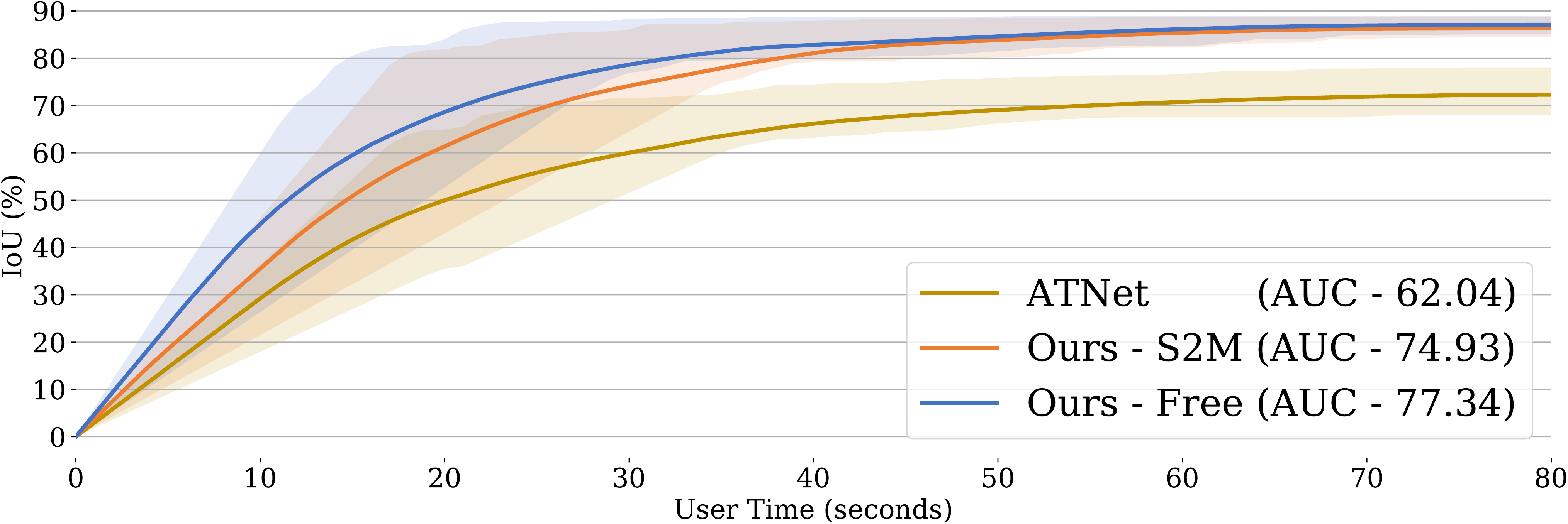}
		\end{center}
		\vspace{-0.2in}
		\caption{
			Mean IoU versus user time plot with shaded regions showing the interquartile range. Our methods achieve higher final accuracy and AUC than ATNet~\cite{Yuk2020IVOSGlobalLocal}. In Ours-Free, users make use of f-BRS~\cite{sofiiuk2020fbrs} to obtain a faster initial segmentation. Experienced users can use free hand drawing and local control to achieve  higher final accuracy given more time. 
		}
		\label{fig:user_auc}
		\vspace{-0.2in}
	\end{figure}

	
	We recruited 10~volunteers who were given sufficient time to familiarize themselves with different algorithms and the GUI. They were asked to label 5~videos in the DAVIS 2017 multi-object validation set with satisfactory accuracy as fast as possible, within a 2-minute wall clock time limit. To avoid familiarity bias, they studied the images and ground truths of each video before each session. Figure~\ref{fig:user_auc} shows the IoU versus user-time plot and Table~\ref{tab:user_delta} tabulates the average performance gain after each interaction. Our method achieves better results with less interaction time, while including more interaction options (f-BRS, free-hand drawing, and local control) which allows our method to converge faster and to a higher final accuracy for experienced users.
	
	\begin{table}[h]
		\centering
		\small
		\vspace{-0.05in}
		\begin{tabular}{l|c|c|c|c|c|c}
			\hline
			Methods & $\Delta_1$  & $\Delta_2$  & $\Delta_3$ & $\Delta_4$ & $\Delta_5$ & Sum\\
			\Xhline{3\arrayrulewidth}
			ATNet~\cite{Yuk2020IVOSGlobalLocal} & 62.2 & 6.82 & 1.93 & 2.57 & 1.61 & 75.1 \\
			Ours-S2M & 83.8 & 1.56 & 0.64 & 0.37 & 0.53 & 86.9 \\
			Ours-Free & 84.3 & 1.69 & 0.66 & 0.66 & 0.62 & 87.9 \\
			\hline
		\end{tabular}
		\vspace{0.1in}
		\caption{Mean incremental IoU improvement after each interaction round. $\Delta_i$ denotes the IoU gain after the $i$th frame interaction and propagation. ATNet~\cite{Yuk2020IVOSGlobalLocal} requires more interactions to achieve stable performance while ours achieves higher accuracy with less interactions.
			Enabling other interaction modes such as f-BRS or local control (Ours-Free) is beneficial to both the speed and the final accuracy. Note that sum does not equal to the final mean IoU in the left plot because not all users interacted for five rounds.}
		\label{tab:user_delta}
		\vspace{-0.15in}
	\end{table}
	\section{Conclusion}
	We propose MiVOS, a novel decoupled approach consisting of three modules: Interaction-to-Mask, Propagation and Difference-Aware Fusion. 
	By decoupling interaction from  propagation, MiVOS is versatile and not limited by the type of interactions. On the other hand, the proposed fusion module reconciles interaction and propagation by faithfully capturing the user's intent and mitigates the information lost in the decoupling process, thus enabling MiVOS to be both accurate and efficient.
	We hope our MiVOS can inspire and spark future research in iVOS.
	
	\newpage
	
	{\small
		\bibliographystyle{ieee_fullname}
		\bibliography{mivos}
	}
	
\end{document}